%% file: sn-article.tex
\documentclass[pdflatex,sn-nature]{sn-jnl}% Style for submissions to Nature Portfolio journals

\usepackage{graphicx}%
\usepackage{multirow}%
\usepackage{amsmath,amssymb,amsfonts}%
\usepackage{amsthm}%
\usepackage{mathrsfs}%
\usepackage[title]{appendix}%
\usepackage{xcolor}%
\usepackage{textcomp}%
\usepackage{manyfoot}%
\usepackage{booktabs}%
\usepackage{algorithm}%
\usepackage{algorithmicx}%
\usepackage{algpseudocode}%
\usepackage{listings}%
\usepackage{multirow}
\usepackage{subcaption}
\usepackage{soul}
\usepackage{makecell}
\usepackage{pifont}
\usepackage{float}
\usepackage{placeins}
\theoremstyle{thmstyleone}%
\theoremstyle{thmstyletwo}%

\theoremstyle{thmstylethree}%

\begin{document}

\title[Article Title]{The Spectrascapes Dataset: Street-view imagery beyond the visible captured using a mobile platform}
% Street view imagery beyond optical photos: 

% Street view imagery beyond the visible: an open multispectral dataset 
%Multispectral street-view imaging data for urban spaces using a generalized mobile platform

% Other Titles

\author[1]{\fnm{Akshit} \sur{Gupta}}\email{a.gupta-5@tudelft.nl}
\author[1]{\fnm{Joris} \sur{Timmermans}}\email{joris.timmermans@tudelft.nl}
\author[2]{\fnm{Filip} \sur{Biljecki}}\email{filip@nus.edu.sg}
\author[1]{\fnm{Remko} \sur{Uijlenhoet}}\email{r.uijlenhoet@tudelft.nl}



\affil[1]{\orgname{TU Delft}, \orgaddress{\street{Stevinweg 1}, \city{Delft}, \postcode{2628CN}, \state{Zuid-Holland}, \country{The Netherlands}}}
% \affil[2]{\orgdiv{CiTG faculty}, \orgname{TU Delft}, \orgaddress{\street{Stevinweg 1}, \city{Delft}, \postcode{2628CN}, \state{Zuid-Holland}, \country{The Netherlands}}}

\affil[2]{\orgname{National University of Singapore}, \orgaddress{\street{4 Architecture Dr}, \city{Singapore}, \postcode{117566}, \country{Singapore}}}

% \affil[2]{\orgdiv{Department}, \orgname{Organization}, \orgaddress{\street{Street}, \city{City}, \postcode{10587}, \state{State}, \country{Country}}}

% \affil[3]{\orgdiv{Department}, \orgname{Organization}, \orgaddress{\street{Street}, \city{City}, \postcode{610101}, \state{State}, \country{Country}}}
\include{sections/abstract}

\maketitle

\input{sections/backgroundAndSummary}
\input{sections/Methods}

\input{sections/technicalvalidation}
\input{sections/datarecords}
\input{sections/usagenotes}

\input{sections/codeAvailablity}

\bmhead{Acknowledgments}
We would like to thank Jaap Elstgeest from the TU Delft DEMO for helping Akshit Gupta in developing the hardware casing. We would like to thank the TU Delft Climate Action Seed Fund for funding the hardware components. The high-resolution Pléiades NEO satellite imagery data was provided by the European Space Agency (ESA) under the Third Party Missions (TPM) program (Project PP0103086); \textcopyright\ Airbus Defence and Space SAS 2025.

\bmhead{Author Contributions}
Akshit Gupta: Conceptualization, Software, Hardware, Methodology, Investigation, Formal analysis, Data Curation, Visualization, Writing- original draft; Joris Timmermans: Methodology, Writing- original draft, Supervision; Filip Biljecki: Writing- review and editing, Visualization, Supervision; Remko Uijlenhoet: Methodology, Writing- review and editing, Supervision, Project administration.

\bmhead{Corresponding author}
Correspondence to Akshit Gupta.

\bibliography{sn-bibliography}% common bib file

\end{document}

%% file: sections/abstract.tex
\abstract{
High-resolution data in spatial and temporal contexts is imperative for developing climate resilient cities. Current datasets for monitoring urban parameters are developed primarily using manual inspections, embedded-sensing, remote sensing, or standard street-view imagery (RGB). These methods and datasets are often constrained respectively by poor scalability, inconsistent spatio-temporal resolutions, overhead views or low spectral information. We present a novel method and its open implementation: a multi-spectral terrestrial-view dataset that circumvents these limitations. This dataset consists of 17,718 street level multi-spectral images captured with RGB, Near-infrared, and Thermal imaging sensors on bikes, across diverse urban morphologies (village, town, small city, and big urban area) in the Netherlands. Strict emphasis is put on data calibration and quality while also providing the details of our data collection methodology (including the hardware and software details). To the best of our knowledge, Spectrascapes is the first open-access dataset of its kind. Finally, we demonstrate two downstream use-cases enabled using this dataset and provide potential research directions in the machine learning, urban planning and remote sensing domains.}

%% file: sections/backgroundAndSummary.tex
\section*{Background \& Summary}
\label{sec:background}

Urban environmental monitoring is a fundamental requirement for building climate adaptive cities \cite{IPCC}. By 2050, it is expected that over two-thirds of the world's population will live in urban areas \cite{IPCC}. As these cities densify, they modify their own local infrastructure, microclimates and environments at ever increasing pace \cite{densification1, densification2}, in addition to the impact of global climate change. Hence, it is imperative to improve the monitoring of urban environments at high temporal and spatial resolutions, to assess both the progress of climate adaptation and evaluate the effects of new policies in uncertain situations \cite{IPCC}.

Nowadays, the common approaches for monitoring the various urban environmental parameters (e.g., buildings, trees, roads etc.) comprise in-person field surveys (field-based monitoring), embedded sensors at particular locations (in-situ monitoring), or remote sensing imagery using satellites and airborne platforms. All these approaches entail various challenges \cite{gupta_2024_tools}. The in-person visits generate high-quality data at high costs, thus making them non-scalable to large urban areas. Similarly, embedded sensors generate data at high-temporal resolution, but with limited spatial coverage. In contrast, remote sensing generates massive amounts of imagery data with high spatial coverage, but can only capture the overhead view of the urban canopy layer. 
% For low-earth-orbit satellites, the temporal resolution over a given location is also limited. 
Furthermore, the spatial granularity of these remote sensing datasets is limited by the imaging sensor deployed on the satellite or the airborne platform \cite{gupta_2024_tools}. In response to these challenges, additional technology-assisted methods have been developed in recent years, to complement these approaches and circumvent their limitations. These methods comprise mobile sensing, which includes specialized sensors deployed on moving vehicles \cite{desouza_2020_air}, or citizen-science based opportunistic monitoring \cite{citizenScience}, which includes utilizing the sensors on smartphones \cite{opportunisticSensingAmsterdam}. 

Specifically in the field of imagery data, researchers are increasingly utilizing street-view imagery data from various platforms such as Google Street View (GSV), Mapillary, KartaView and Baidu Maps \cite{biljecki_2021_street}. These efforts usually apply computer vision methods to inspect urban environmental parameters of interest. They span across a wide range of urban parameters, from using GSV images to developing a catalog of urban greenery \cite{BRANSON201813}, detecting the material of façades \cite{tarkhan2025mapping}, to using them along with other aggregated data to assess bikeability of streets \cite{2021_trc_bikeability}.

Both street view based methods along with satellite and airborne platforms have been deemed scalable enough for entire cities \cite{DENG2021125}. However, street-view imagery based datasets are largely limited by the spectral information in the imagery data (usually only Red, Green, Blue (RGB)), while providing terrestrial view of the urban canopy where humans work and live. For instance, one study \cite{2024_global_streetscapes} bundles millions of RGB street-view images from the Mapillary and Kartaview platforms spanning across diverse cities and enhances them with labels. However, the limited number and range of spectral bands make it difficult to evaluate the suitability of the images for specific analyses with environmental indices \cite{2024_global_streetscapes}, leading to qualitative biases, rather than quantitative evaluations based on calibrated physical indices \cite{2026_land_greenery}. On the other hand, satellite and airborne-based datasets are limited in terms of overhead view of the urban canopy, although they provide imagery data with multiple spectral bands in addition to the Red, Green and Blue \cite{Loudjani2023Pleiades}. In line with this, recent projects have investigated gathering street-view imagery in addition to Red, Green and Blue spectral bands using custom hardware deployments. For instance, Greenscan \cite{greenscan} utilizes thermal and multispectral imaging sensors deployed on moving vehicles to detect the health condition of trees, albeit in a limited area, while another study \cite{instantInfrared} utilizes only thermal imaging sensors to detect energy leakage from façades. However, these works focus on specific areas, or spectral bands with specific sensors, limiting the scope of these studies. Further, the datasets gathered are sometimes not available openly. Table \ref{tab:comparison} shows the properties of these datasets.

\begin{table}[h]
\centering
\resizebox{\textwidth}{!}{%
\begin{tabular}{lcccccl}
\toprule
Work & Morphological Diversity & \multicolumn{3}{c}{Spectral Bands in Imagery} & Custom Platform & Open Access\\
\cmidrule(lr){3-5}
 &  & RGB & NIR & Thermal &  \\
\midrule
\cite{instantInfrared}     & Limited (1 city)         & \ding{55} & \ding{55} & \ding{51} & No & No  \\
\cite{greenscan}                 & Limited (1 city)         & \ding{51} & \ding{51} & \ding{55} & Yes & No  \\
\cite{tarkhan2025mapping}        & Limited (3 metro.\ hubs) & \ding{51} & \ding{55} & \ding{55} & No & Yes \\
\cite{2024_global_streetscapes}  & Diverse      & \ding{51} & \ding{55} & \ding{55} & No & Yes \\
\textbf{This work}               & Diverse (4 morphologies) & \ding{51} & \ding{51} & \ding{51} & Yes & Yes \\
\bottomrule
\end{tabular}%
}
\caption{Comparison of recent street-view imagery datasets by spatial coverage, spectral richness, and open accessibility, and our contribution.}
\label{tab:comparison}
\end{table}

In this work, we provide an open dataset called Spectrascapes which consists of calibrated Red, Green, Blue, Near-Infrared and Thermal imaging bands from street-view level across four diverse urban morphologies in the Netherlands. We also provide the coordinates where these multi-spectral images were captured, which allows combination with very-high resolution (30 cm/pixel) satellite imagery, such as Pleiades NEO \cite{Airbus2021PleiadesNeo} with multi-spectral bands. As a result, this work provides comprehensive multi-spectral street-view imaging data with metadata for comprehensive terrestrial and overhead satellite imagery across diverse urban morphologies. The Spectrascapes dataset provides three main contributions:
\begin{enumerate}
    \item Spectral synergy: This dataset enables the calculation of physics-based indices from the street-view level, such as Normalised Difference Vegetation Index (NDVI), Land Surface Temperature (LST) and energy leakage of façades. These were previously out of reach using the standard street view platforms with only RGB spectral bands, such as Google Street View or Mapillary.
    \item Cross-scale validation potential: Using the metadata provided, the readers can obtain Very-High Resolution (VHR) overhead satellite imagery, such as Pleiades NEO. Thus, this work acts as a bridge to ensure that any method developed using this dataset can be validated on scale with VHR satellite-based imagery.
    \item Diversity, quality, and transparency: The four different urban areas covered have different urban morphologies. Furthermore, as shown in Section \texttt{Technical Validation}, the quality of our images is robust with multiple calibrations and data processing. We also provide the details and code for our hardware collection device, encouraging researchers to reproduce and contribute more data with different spectral bands to this dataset. 
\end{enumerate}

\begin{figure}[ht]
\centering
\includegraphics[width=\linewidth]{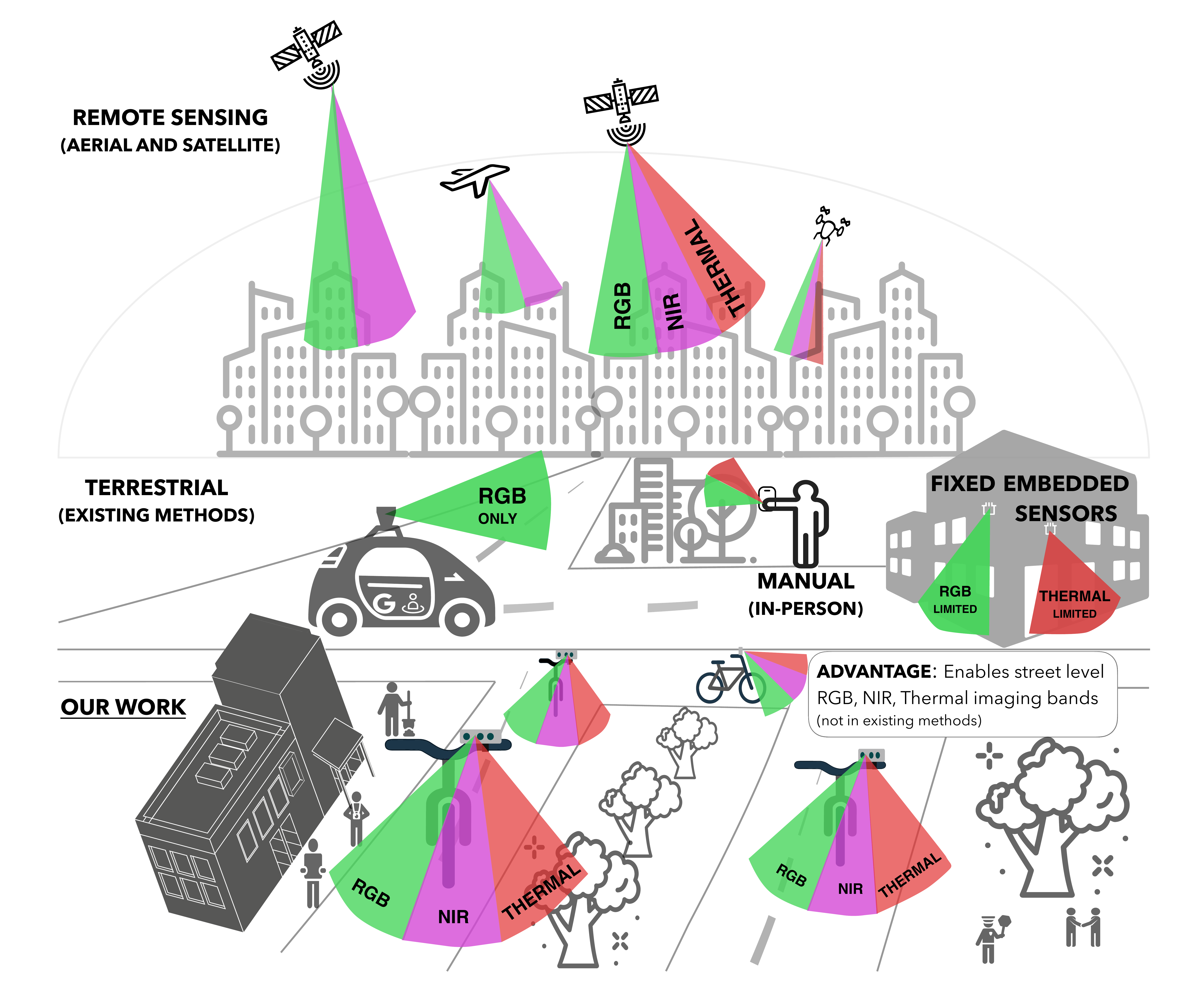}
\caption{Overall summary of our approach for data collection, using bikes to capture multi-spectral terrestrial imagery data.}
\label{fig:overallSummary} 
\end{figure}

Figure \ref{fig:overallSummary} illustrates our work and situates it within the context of existing approaches. By offering multiple spectral bands in the images associated with this dataset, this work enables a synchronous multi-task setting for various experimental scenarios involving multiple urban physical indices. Further, this dataset enables various applications from street-view level, including pretraining on synchronized spectral bands to generate spectral bands where only RGB imagery is available for the machine learning field, as well as urban greenery mapping and mapping of building parameters for the urban planning field.

We first present the details about our mobile platform used to collect imagery data, and how it can be extended for more applications. Then, we present the validated imagery data with calibration focused mainly on the Red, Green, Blue, Near-Infrared and Thermal spectral bands in the diverse urban morphologies. Lastly, we demonstrate certain use cases that the Spectrascapes dataset enables, along with multiple research directions, directly emanating from it and our flexible data collection approach.

%% file: sections/Methods.tex
\section*{Methods}
\label{sec:methodology}

The imagery data collected in this work was collected by mounting custom-made hardware on a moving bike (bicycle). The data was captured autonomously at fixed time intervals while the bike traversed narrow, segregated bike lanes in the Netherlands. In this section, we will expand on the hardware design, its architecture and the data collection campaigns across the four urban morphologies.

\subsection*{Hardware Design}

\begin{figure}[h]
\centering
\includegraphics[width=\linewidth]{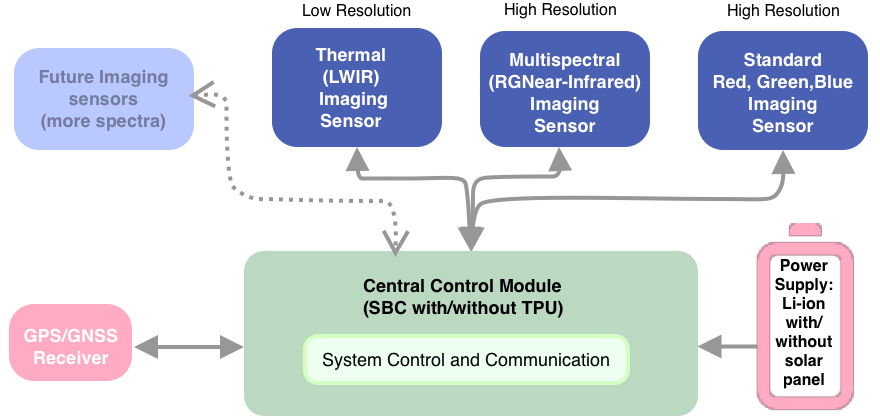}
\caption{Architecture of our system. Refer to Table \ref{tab:hardware} for details of each imaging sensor.}
\label{fig:architecture} 
\end{figure}

Our hardware is custom-designed based on the imagery collection hardware of recent works, mainly the GreenScan platform \cite{greenscan}. However, while Greenscan \cite{greenscan} focuses only on measuring two physics-based indices for urban trees (i.e., the Normalized Difference Vegetation Index (NDVI), and the Canopy Temperature Difference (CTD)), our architecture is designed to be flexible for capturing imagery data consisting of multiple spectral bands. Flexibility of capturing imagery data from multiple spectral bands was our top priority to enable monitoring of multiple urban environmental parameters. Further, as explained in the previous Section \texttt{Background and Summary}, an additional requirement was the suitability of the hardware deployment for mobile sensing to ensure scalability, either using bikes (bicycles), or opportunistic environmental monitoring by citizens. Unlike automobiles, bikes have specific advantages, such as the ability to enter narrow streets in the city centres with their increased maneuverability due to reduced speeds. In addition, the bike movement trajectories are closer to the built environment, thus generating spatial imagery with increased details. The architecture of the custom-developed hardware is shown in Figure \ref{fig:architecture}.
 
\begin{figure}[h]
\centering
  \begin{subfigure}[t]{0.5\linewidth}
    \centering
    \includegraphics[width=\linewidth]{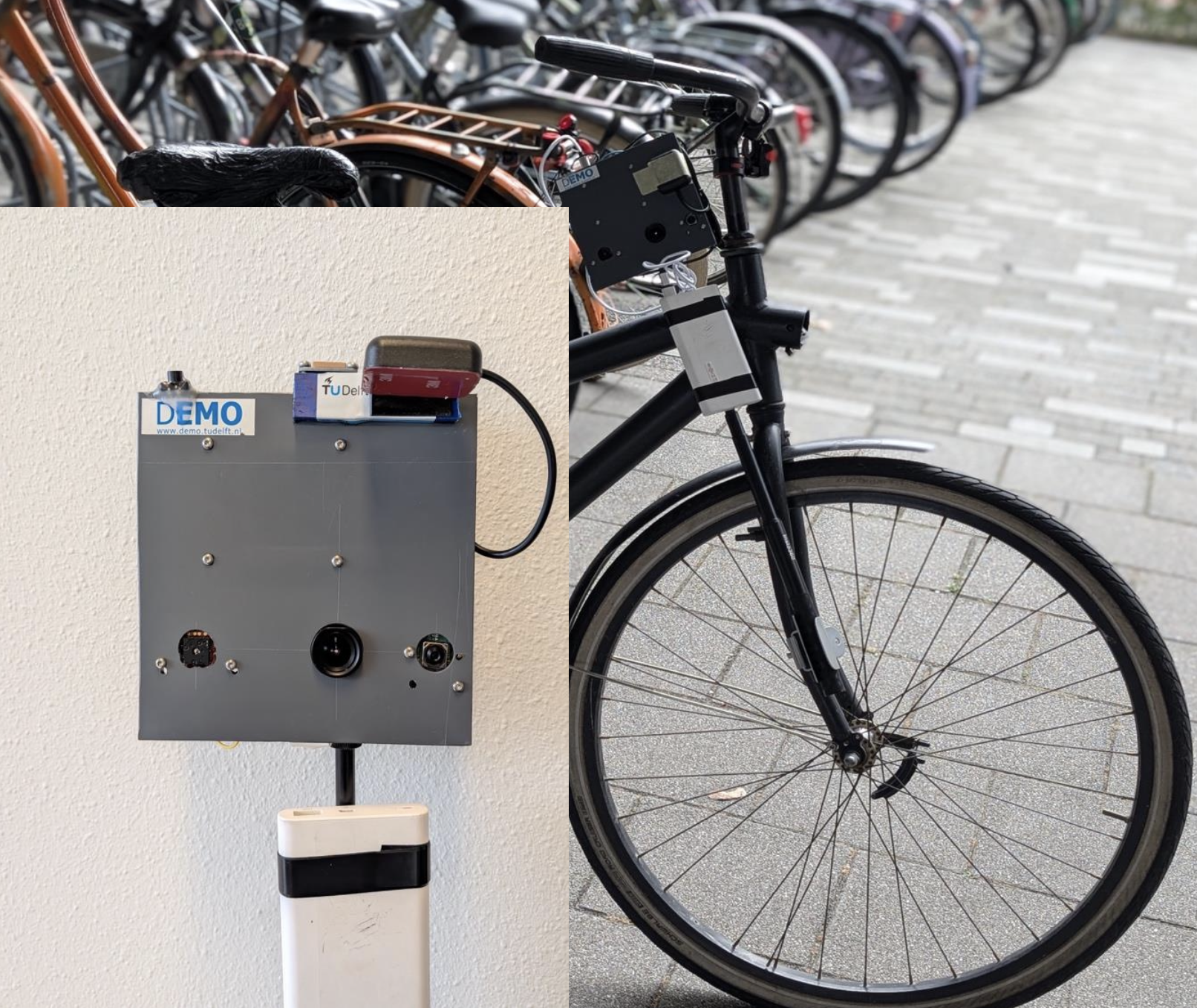}
    \caption{The hardware with all the three imaging sensors, control unit and battery.}
  \end{subfigure}
   \begin{subfigure}[t]{0.35\linewidth}
    \centering
    \includegraphics[width=\linewidth]{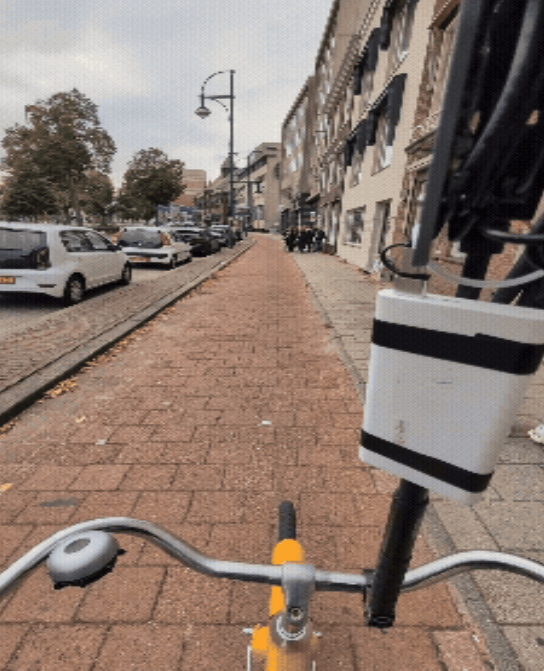} 
    \caption{The device attached to a rental bike.}
  \end{subfigure}%%
\caption{Hardware and its deployment on a rental bike (OV-fiets bicycle) in the Netherlands.}
\label{fig:platformAndBike} 
\end{figure}

Each hardware platform has three imaging sensors (i.e, a Red, Green, Blue (RGB) imaging sensor, a Red, Green, Near-Infrared (RGN) imaging sensor and a Thermal (Long-Wave Infrared) imaging sensor), which together capture imagery data across five spectral bands. The details of each of these sensors are shown in Table \ref{tab:hardware}. Further, the platform allows the integration of higher cost imaging sensors to capture additional spectral bands if needed. A central control and computing platform (Raspberry Pi 3 in this case) enables the integration of these imaging sensors, while allowing the storage of data. The shutter speeds in the imaging sensors were set to the best available setting with the lowest ISO setting for reducing the possibility of blurry images, while a bike is moving. The synchronized image capture for the multiple imaging sensors is achieved through connection signals from the central control and computing platform. Due to inherent manufacturer constraints, the RGN sensor actuates with a fixed systematic temporal delay of 0.22 seconds. Figure \ref{fig:platformAndBike} shows the entire hardware platform and its deployment on a moving bike in the Netherlands.

\begin{table}[h]
\centering
\resizebox{\textwidth}{!}{%
\begin{tabular}{lllllll}
\toprule
\makecell[l]{\textbf{Imaging} \\ \textbf{Sensor}} & \textbf{Sensor Name}                                    & \textbf{Spectral Bands} & \textbf{Image Size (pixels)} &\makecell[l]{\textbf{Manufacturer}\\ \textbf{(Accuracy)}} & \makecell[l]{\textbf{Connection Signals}\\ \textbf{(from control unit)}} & \textbf{High-cost substitution}  \\
\midrule
RGN            & Mapir Survey 3W RGN \cite{MapirSurvey3W}                           & \makecell[l]{500-575nm (Green)\\630-690nm (Red)\\810-890nm (N-Infrared)} & 4000x3000 & $<1\%$ (Lens)   & \makecell[l]{PWM} & Micasense Altum-PT \cite{MicaSenseAltumPT} \\
\midrule
Thermal        & Flir Lepton 3.5 \cite{FlirLepton35}                               & \makecell[l]{8,000nm-14,000nm\\(Long-Wave Infrared)}                         & 160x120 & $\pm 5^{\circ}\text{C}$ or $5\%$   & RPC over UART   & Flir Boson \cite{FlirBoson640}             \\
\midrule
RGB            & \makecell[l]{Pi Camera Module 3\\Sony IMX708}~\cite{RaspberryPiCamera3}  & \makecell[l]{480–600nm (Green)\\590-700nm (Red)\\400-540nm (Blue)} & 4608×2592  & N/A & I2C and MIPI CSI-2  & Multiple Available                            \\
\bottomrule
\end{tabular}%
}
\caption{Details of the imaging sensors utilized in the hardware platform.}
\label{tab:hardware}
\end{table}

\subsection*{Data Collection Campaigns}

\begin{figure}[h]
\centering
\includegraphics[width=\textwidth]{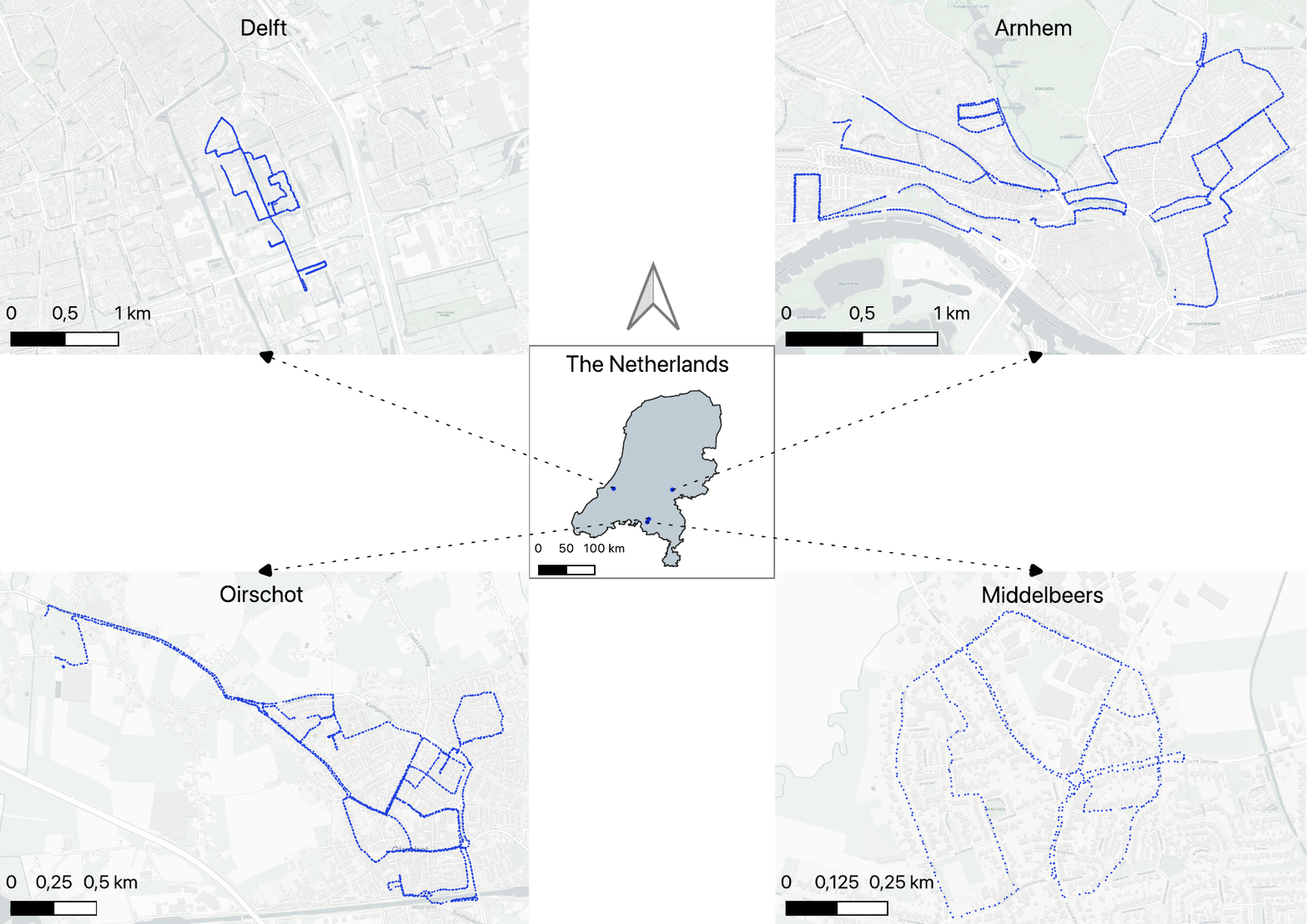}
\caption{The study areas in the Netherlands. The blue dots indicate the locations where the multi-spectral imaging data was gathered with a bicycle. Arnhem is a large city with a population of around 167,632 people, Delft is a small city with a population of around 109,577 people, Oirschot is a town with a population of 19,280 people and Middelbeers is a small village with a population of 4,994 people. All maps presented with a north-up orientation. Population numbers sourced from \cite{cbs_keyfigures}.}
\label{fig:locationOfDataObservation}
\end{figure}

We gathered data in four different urban morphologies in the Netherlands using the hardware designed from bottom-up. These study areas include Arnhem, Delft, Oirschot and Middelbeers, i.e. a big urban area, a small urban city (focusing on the university campus), a town and a village. These were strategically selected based on different urban parameters, such as the population of the urban area, the elevation variation in the urban area and their geographical location in the Netherlands (east to west), to capture a comprehensive cross-section of Dutch urban morphologies. The four study areas are distributed across a longitudinal gradient, capturing the geological and topographical diversity of the Netherlands. This gradient extends from the low-lying western coastal region, exemplified by Delft situated below mean sea level on peat and clay soil, to the elevated eastern and southern regions with more sandy soils. The latter is represented by Arnhem, a topographically varied city reaching elevations of 90 m above mean sea level at about 30 km from the German border. Complementing these are Oirschot and Middelbeers, located within a predominantly flat, rural agricultural landscape in the southern interior, approximately 15 km west of the city of Eindhoven. Together, these four sites encapsulate the demographic, geological, and elevation parameters in the country. The locations of these study areas are shown in Figure \ref{fig:locationOfDataObservation}.

The imagery data collection is further facilitated by the bike lanes present in almost all the parts of the Netherlands, even though the underlying platform is agnostic to bike lane availability. We collected data in the months of September and October 2025, taking almost three days in each urban area for imagery data collection. The total biking distance for data collection was approximately 82 km. During these three days, multiple trajectories were followed with suitable breaks in between. 

Given the opportunistic nature of naturalistic biking trajectories, some streets have been covered more than once. Further, it was ensured that the weather was cloudy during data collection to allow soft diffuse illumination from the sun. Additionally, to ensure that the sun is at or close to nadir position, the data collection was performed between the time interval of 11 am to 3 pm (local time) each day. In Delft and Arnhem, the device was always facing the right side of the direction of motion of the bicycle at an angle close to ninety degrees, which was possible due to the wide streets. In Oirschot and Middelbeers, on the other hand, due to the small width of the streets, to increase the field of view of the imagery, the device was always facing the left hand side of the direction of motion of the bicycle. Figure \ref{fig:locationOfDataObservation} shows the locations where the imaging data was captured in each of the four urban areas.

\begin{figure}[htbp]
     \centering
     % Top Left
     \begin{subfigure}[b]{0.45\textwidth}
         \centering
         \includegraphics[width=\textwidth]{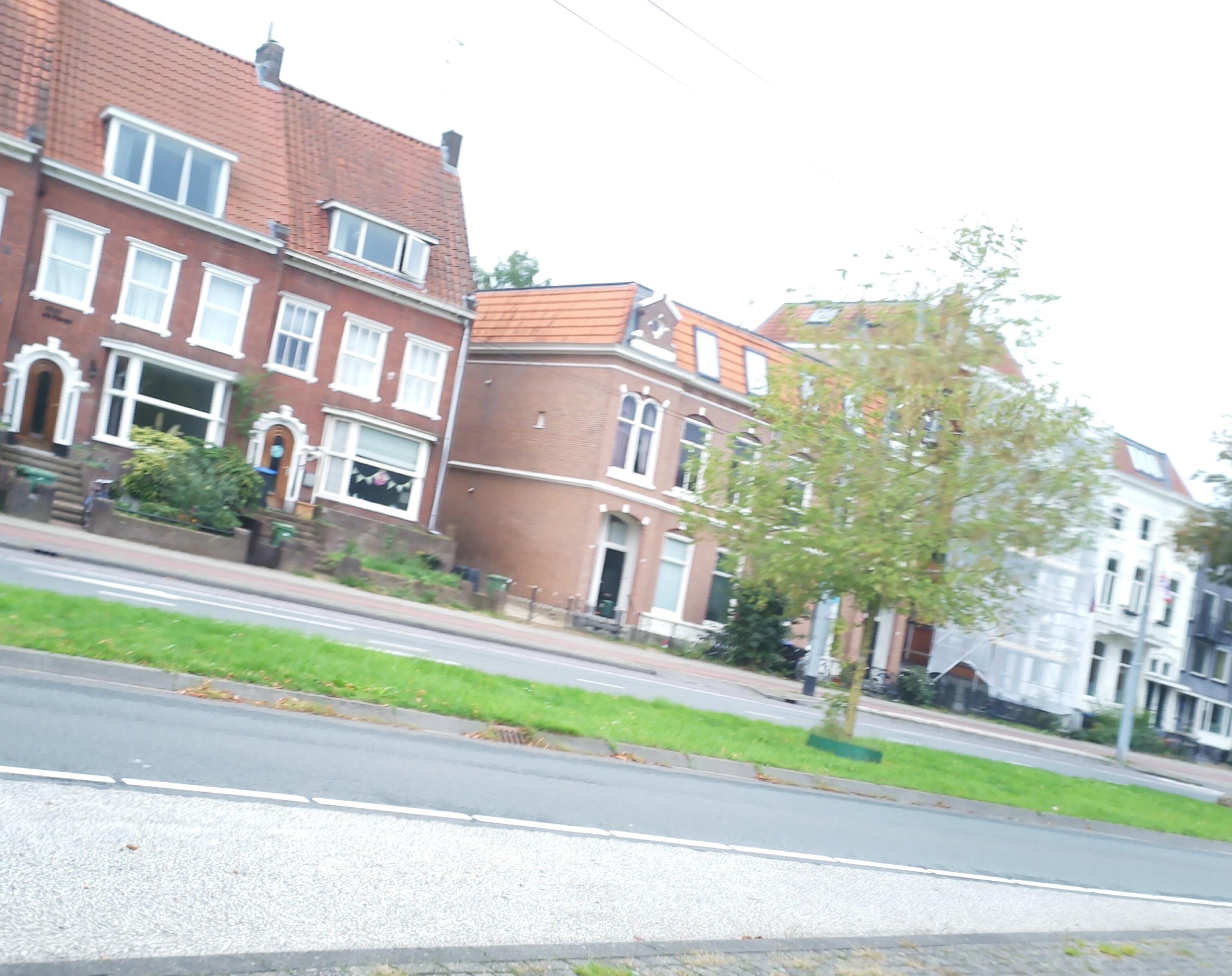}
         \caption{RGB imaging data.}
     \end{subfigure}
     \hfill
     % Top Right
     \begin{subfigure}[b]{0.45\textwidth}
         \centering
         \includegraphics[width=\textwidth]{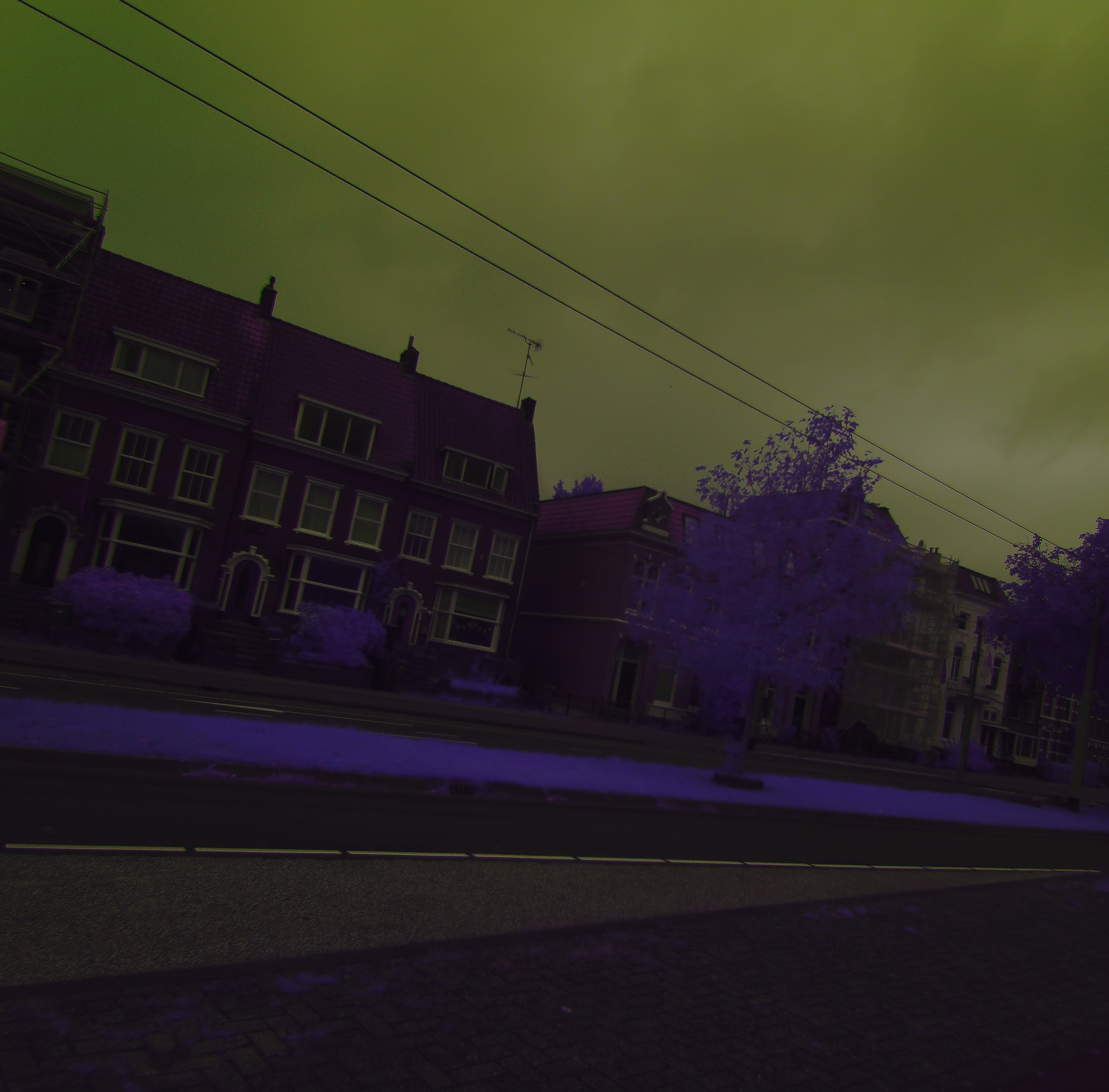}
         \caption{RGNear-Infrared imaging data.}
     \end{subfigure}

     \vspace{10pt} % Adds vertical spacing between rows
     % Bottom Left
     \begin{subfigure}[t]{0.45\textwidth}
         \centering
         \includegraphics[width=\textwidth]{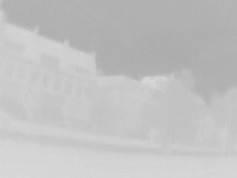}
         \caption{Thermal (LWIR) imaging data.}
     \end{subfigure}
     \hfill
     % Bottom Right
     \begin{subfigure}[t]{0.45\textwidth}
         \centering
         \includegraphics[width=\textwidth]{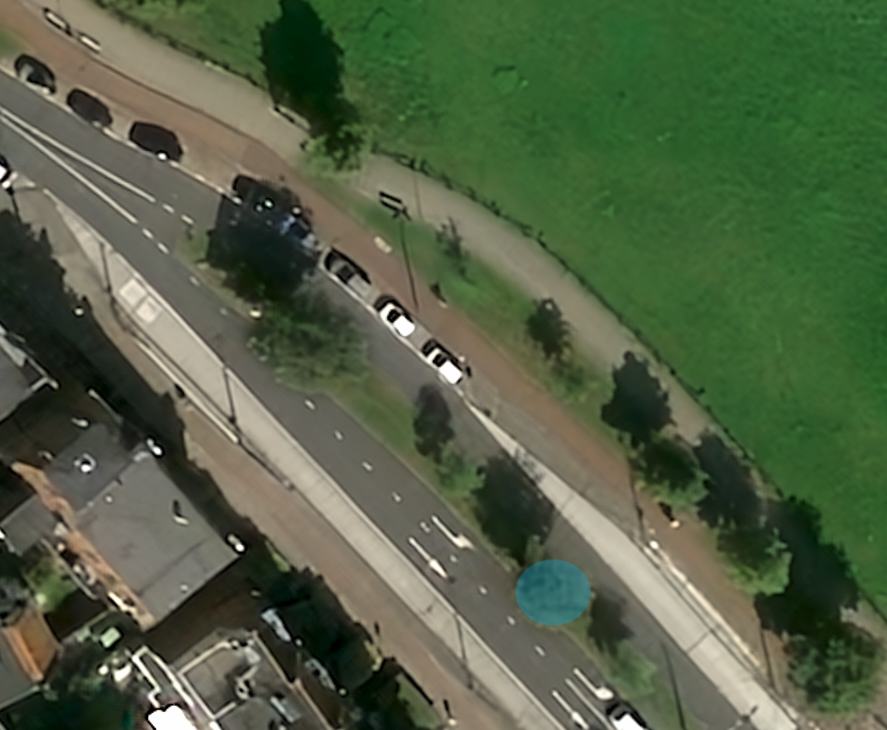}
         \caption{Very-high resolution satellite imagery from Pléiades NEO obtained using the location metadata of the data observation location. \textcopyright\ Airbus Defence and Space SAS 2025, provided by ESA.}
     \end{subfigure}
    \caption{Example of the multi-spectral imaging data, and the remote sensing image obtained using the metadata (Lat: 51.986186667, Long: 5.904403333) at each data observation location. This data was captured on Zijpendaalseweg street in Arnhem.}
    \label{fig:datasinglePoint}
\end{figure}

Each data point in Figure \ref{fig:locationOfDataObservation} contains images from three different sensors as explained earlier. Thus, each single data observation point contains street-level imaging data and associated metadata, which can be leveraged to retrieve corresponding overhead satellite imagery. This is shown in Figure \ref{fig:datasinglePoint}. The Pleiades Neo satellite imagery data, featuring a 30 cm spatial resolution as shown in Figure \ref{fig:datasinglePoint}, was sourced from the ESA Third Party Missions (TPM) program \cite{ESA_PleiadesNeo_TPM}.

%% file: sections/technicalvalidation.tex
\section*{Technical Validation}
\label{sec:validation}

\subsection*{Imagery data collection protocol}

The data presented in this work consists of three distinct modalities of images: street-level RGB, street-level RGN and street-level thermal images, each gathered with its own sensor from a moving bicycle. 

During data collection, the sensors were all measuring either the reflected or thermally emmitted radiance. Due to dynamic variability, such as cloud cover and the solar position in the sky, the amount of irradiance received by the urban canopy and transmitted to the sensors is also variable at a particular time. This means that the amount of energy (or irradiance) reaching the earth's surface from the remote point source, i.e. the sun, is dependent upon the presence, distribution and density of clouds, and various other constituents of the atmosphere.

When transforming the measured radiance into a meaningful metric (such as surface reflectance, surface temperature), a distinction needs to be made between two types of reflectances: the directional reflectance and the diffuse reflectance. Specifically, the irradiance illuminating the surface has both a diffuse sky component (whose intensity is uniform across all directions) as well as a directional solar component (depending on the solar angle). Further, as our hardware collection device is not static, additional directionality is introduced due to the objects being illuminated by reflected irradiance from adjacent objects such as buildings, cars etc. \cite{USTIN1990293}. In such a case, as much as 20 percent of the illumination in the 750 nm - 1200 nm range can be attributed to sunlight (irradiance) scattered off the surrounding objects \cite{USTIN1990293}. Furthermore, objects in the surrounding area also obscure portions of the diffuse skylight and thereby provide additional directional effects. In an urban environment, this leads to significant challenges, as this directionality depends on the location of the object of interest on a particular street. For example, performing observations on the sidewalk next to a tall office building (with a high degree of glass) will give rise to a large building-reflected sunlight (irradiance) contribution. In contrast, performing the same observation in the middle of the street will significantly lower this contribution.

Based on this, the total irradiance of an object can be calculated as:
\begin{equation}
E _{\textrm{tot}}(\theta_s, \phi_s)=E_{d} + E_{s} (\theta_s,\phi_s) + \sum_{\textrm{obj}}(\rho_{d}^\textrm{obj} * E_{d} + \rho_s^\textrm{obj} * E_{s} (\theta_s,\phi_s))
\label{eq:reflactance}
\end{equation}
where:
\begin{itemize}
     \item \(E_{\textrm{tot}}\) represents the total irradiance of an object (in \(\mathrm{W/m^2}\))
     \item \(E_{d}\) is the diffuse sky irradiance
     \item \(E_{s}\) is the direct irradiance from the sun for a specific sun zenith \(\theta_s\) and azimuth angle \(\phi_s\)
     \item \(\rho^{\textrm{obj}}_{d}\) and \(\rho^\textrm{obj}_s\) represent the diffuse and direct reflectances of the surrounding objects, respectively.
\end{itemize}

Hence, it was chosen to adapt the data collection protocol for making accurate irradiance measurements \cite{EnMAPFieldGuidesTechnicalReport} to reduce the contribution of the directional components $E_{s} (\theta_s,\phi_s)$ and $ \rho_s^\textrm{obj} * E_{s} (\theta_s,\phi_s)$ in the total irradiance of an object. Specifically, the data collection was performed only during cloudy conditions. Without sunlight directly illuminating the object, $E_{s} (\theta_s,\phi_s)=0$ in \eqref{eq:reflactance}. Thus, the total irradiance of an object originates from only from the reflections of diffuse irradiances, \(E_{d}\) and $\rho_{d}^\textrm{obj}* E_{d}$. An additional benefit of acquiring observations under cloudy conditions is that the spectral dependence of the irradiances decrease. Under cloudy conditions, Mie scattering of sunlight becomes more dominant compared to Rayleigh scattering, resulting in a spectrally flat irradiance curve.

\subsection*{Calibration of Imaging sensors and data post-processing}

For capturing accurate imagery data with multiple imaging sensors, rigorous multi-stage calibration was performed before and during the data collection campaigns with the hardware collection device. We performed intrinsic calibration for each of the imaging sensors before deployment. We also performed extrinsic calibration between the imaging sensors enabling downstream image processing. Finally, due to the issues associated with input radiance to the imaging sensors as explained previously, we also performed spectral calibration at regular intervals during data collection for the Red, Green, Near-infrared imagery. The repositories listed in Section \texttt{Code Availability} provide all the computed intrinsic and extrinsic calibration matrices to allow direct multi-spectral re-projection of imagery data for downstream applications.

\begin{figure}[ht]
     \centering 
     % --- First Row: 3 Images ---
     \begin{subfigure}[h]{0.31\textwidth}
         \centering
         \includegraphics[width=\textwidth]{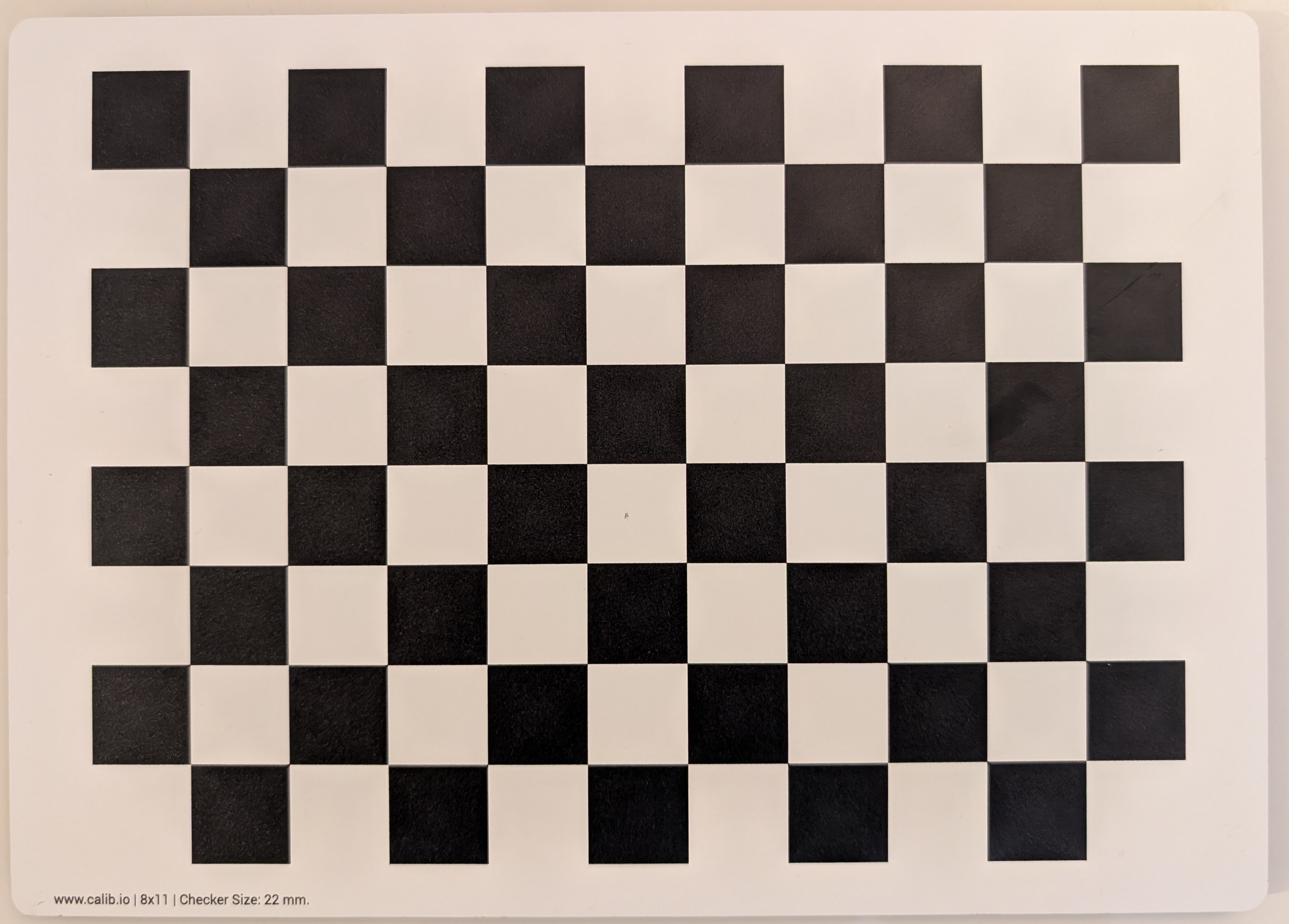}
         \caption{The calibration plate used for intrinsic and extrinsic calibrations of RGB and RGN imaging sensors \cite{calibCheckerboard}. Each side of a square checker measures 22mm. }
         \label{fig:checkerboard}
     \end{subfigure}\hfill
     \begin{subfigure}[h]{0.31\textwidth}
         \centering
         \includegraphics[width=\textwidth]{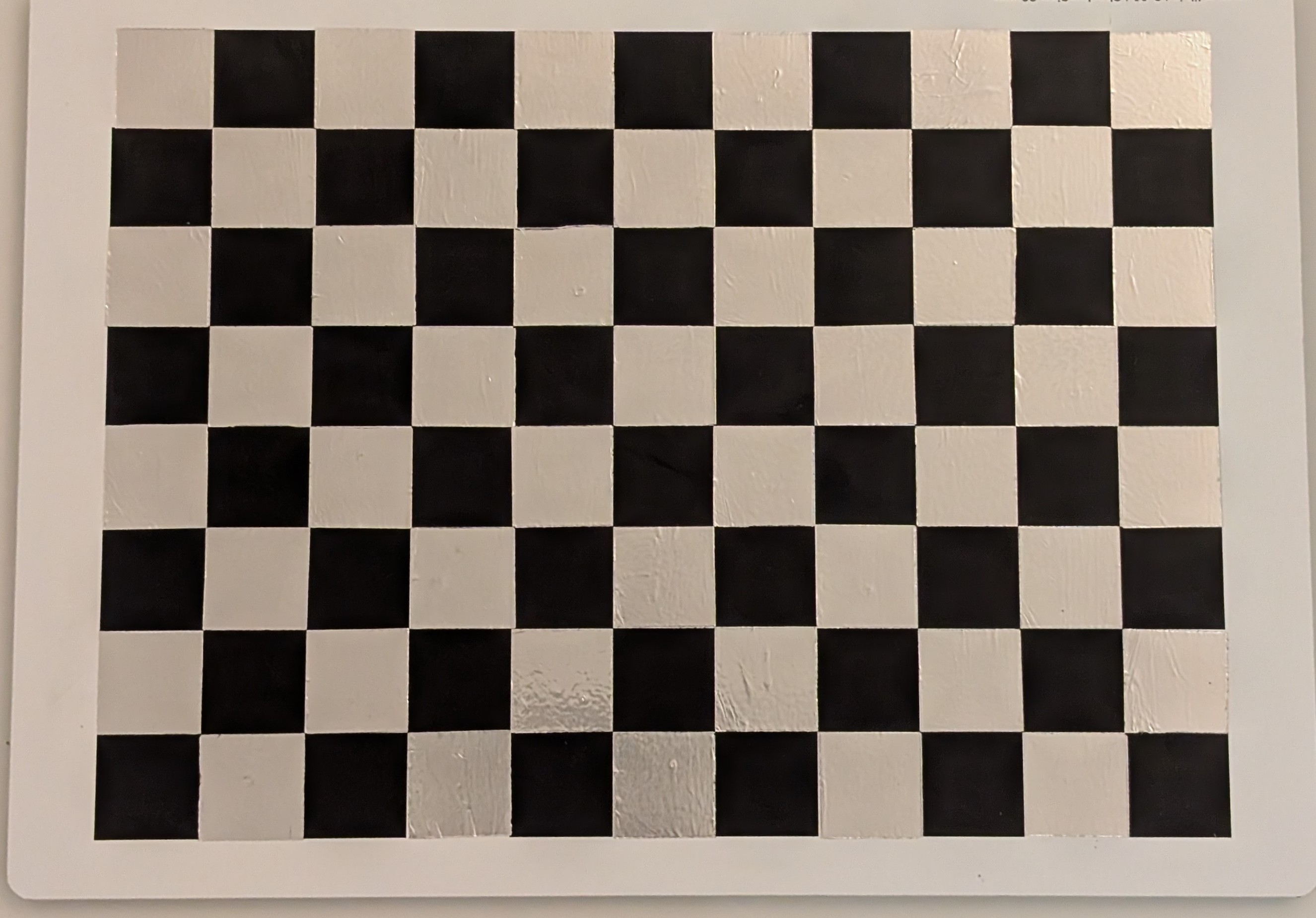}
         \caption{The calibration plate from Figure \ref{fig:checkerboard} modified with the aluminium tape approach from \cite{aluminiumcheckerboard}, to allow intrinsic and extrinsic calibrations of RGB and thermal imaging sensors.}
         \label{fig:alumiumCheckerboard}
     \end{subfigure}\hfill
     \begin{subfigure}[h]{0.31\textwidth}
         \centering
         \includegraphics[width=\textwidth]{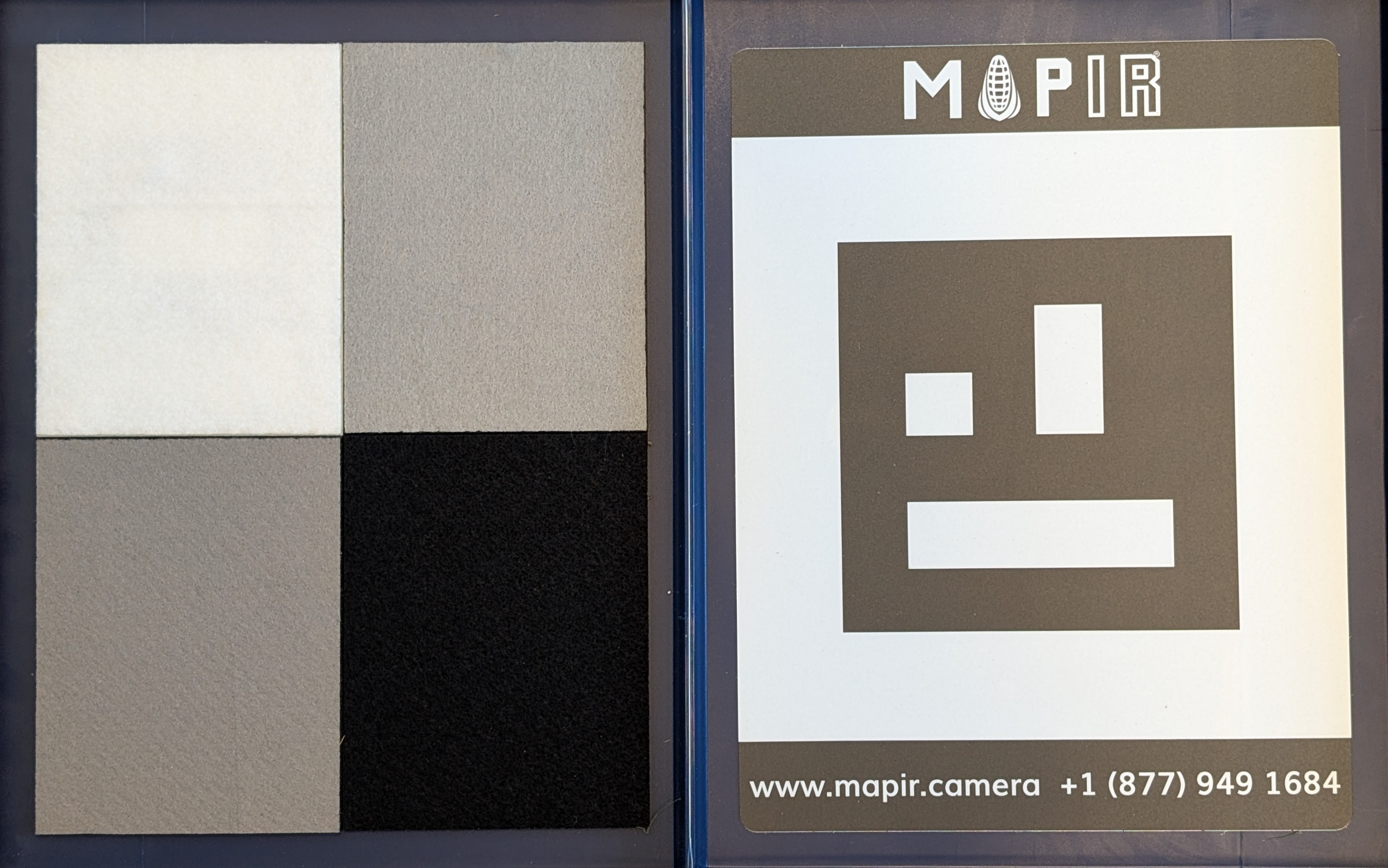}
         \caption{Diffuse reflectance calibration target (V2) from MAPIR \cite{MapirCalibrationPlate} used for spectral calibration of RGN imaging sensor. Each rectangular target of known reflectance measures 139mm x 107mm.}
         \label{fig:mapirCalibrationPlate}
     \end{subfigure}
    \caption{Calibration plates utilized for intrinsic calibration and extrinsic calibration of all the imaging sensors, and spectral calibration of the RGN imaging sensor.}
    \label{fig:calibrationPlates}
\end{figure}

\subsection*{Spectral calibration}
\label{sec:spectralCalbration}

The RGB sensor operated in an uncompensated radiometric state, while the thermal sensor utilized an internal mechanical shutter for spectral self-calibration (Flat-Field-Correction). In contrast, the RGN sensor underwent radiometric calibration via a manufacturer-provided diffuse reflectance standard plate \cite{MapirCalibrationPlate}.

To ensure radiometric consistency and quantitative validity of the RGN data, a radiometric calibration protocol was followed during data collection. The diffuse reflectance calibration target (V2) from MAPIR \cite{MapirCalibrationPlate} as shown in Figure \ref{fig:mapirCalibrationPlate} was utilized to provide known diffuse reflectance standards $(83\%, 27\%, 21\%, 2\%)$ across the visible and near-infrared spectral bands. As explained previously, the illumination conditions, such as the solar position and the cloud cover fluctuate continuously outdoors. However, under the assumption that the distribution of objects per street does not change, the directionality of the total irradiance received from surrounding objects on a particular street to the object of interest does not change either, that is, $\sum_{\textrm{obj}}(\rho_{d}^\textrm{obj} * E_{d})= K$ constant. This reduces \eqref{eq:reflactance} to $E _{\textrm{tot}}=E_{d} + K$ as $E_{s} (\theta_s,\phi_s)=0$ under cloudy conditions.

To compensate for the effect of changing $E_{d}$, raw images of the MAPIR calibration target were captured at strict 30-minute intervals during the data collection process. These reference plate images with known reflectance standards allow the transformation of the raw sensor Digital Numbers (DN) captured in the RAW files into radiometrically corrected surface reflectance values in the RGN images after data post-processing. This spectral calibration for surface reflectances ensures that any spectral index derived from the RGN imagery remains consistent, comparable, and radiometrically valid throughout the dataset.

\subsubsection*{Intrinsic Calibration}

Intrinsic calibration was performed independently for each imaging sensor. A calibration target with white and black checkerboard pattern (8x11) printed using UV-ink on an aluminium plate with checker size of 22 mm as shown in Figure \ref{fig:checkerboard} was utilized to ensure feature extraction (checkerboard corners) for the RGB and RGN imaging sensors. For the visibility of the same features in the thermal spectral band, the calibration target was modified by applying aluminum tape on the alternating squares as shown in Figure \ref{fig:alumiumCheckerboard}. The different emissivities of the aluminium tape and the black UV-ink on the aluminium plate generates a high contrast radiometric pattern visible in both the thermal and RGB spectral bands when placed outdoors in an open field. The intrinsic matrices and distortion coefficients for the imaging sensors were subsequently measured using Zhang's method \cite{checkerboardCalibration} and implemented via the OpenCV framework \cite{opencv_library}. This process allows the accurate modeling of the focal length, optical centers and lens distortion coefficients for each imaging sensor. The Root Mean square (RMS) re-projection error in pixels after intrinsic calibration is shown in Table \ref{tab:calibration_rms}.

\subsubsection*{Extrinsic Calibration}
After intrinsic calibration, the physical translation and rotation between the imaging planes of the imaging sensors was measured using extrinsic calibration. The RGB sensor was designated as the global origin $(0,0,0)$. In other words, to estimate the physical geometry between the imaging sensors, specifically the rotation matrix $R$ and translation matrix $T$ between the RGB imaging sensor, and the thermal and RGN imaging sensors, simultaneous image pairs of the calibration targets (Figures \ref{fig:checkerboard} and \ref{fig:alumiumCheckerboard}) were captured. The radiometrically visible checkerboard corners in their respective calibration targets were extracted across multiple different poses of the checkerboard, including rotation and depth variations. Using the checkerboard corners visible in 2D images and the known 3D geometries, we utilized the stereo calibration algorithm implemented in the OpenCV framework \cite{opencv_library} to optimize the extrinsic parameters. This entire process allows the understanding of the physical geometry of our hardware collection device with the three imaging sensors, thus, enabling highly accurate multi-spectral re-projection of different images. To ensure a unified 3D coordinate system, the intrinsic parameters of the RGB sensor were locked during both the extrinsic stereo calibration routines (RGB $\leftrightarrow$ Thermal and RGB $\leftrightarrow$ RGNear-Infrared), forcing the optimization algorithm to exclusively solve for the spatial transformations ($R$ and $T$). The Root Mean Square (RMS) re-projection errors for the RGB $\leftrightarrow$ Thermal and RGB $\leftrightarrow$ RGN sensor pairs are also shown in Table \ref{tab:calibration_rms}.

\begin{table}[htpb]
\centering
\begin{tabular}{@{}llc@{}}
\toprule
\textbf{Calibration Type} & \textbf{Individual Sensor / Sensor Pair} & \textbf{RMS Error (pixels)} \\ \midrule
\textbf{Intrinsic}        & RGB                    & 0.49                        \\
                          & Thermal                 & 0.35                        \\
                          & RGNear-Infrared                  & 0.58                        \\ \addlinespace
\textbf{Extrinsic (Static)} & RGB $\leftrightarrow$ Thermal & 0.61                        \\
                          &  RGB $\leftrightarrow$ RGNear-Infrared  & 0.54                       \\ \bottomrule
\end{tabular}
\caption{Intrinsic and extrinsic calibration re-projection errors for the imaging sensors to understand the internal characteristics of the imaging sensors and estimating the physical geometry of our data collection hardware respectively.}
\label{tab:calibration_rms}
\end{table}

As detailed in Table \ref{tab:calibration_rms}, across all independent intrinsic and extrinsic calibrations, the RMS re-projection errors are below a single pixel threshold. Given the differences in the resolutions of our imaging sensors ($4000\times3000$ for RGN, $4608\times2592$ for RGB, $160\times120$ for Thermal) and complexities associated with feature matching, these RMS values demonstrate robust geometric correction. This ensures that a physical understanding using intrinsic and extrinsic matrices is avaialable in a digital files, providing a foundation for users of this dataset to pursue various downstream applications depending on their object of interest.

\subsection*{Data post-processing}

After data collection campaigns, we used Egoblur model \cite{raina2023egoblur} to anonymize the dataset, systematically removing any Personally Identifiable Information (PII) such as discernible faces and vehicle license plates. This automated process was supplemented with randomized manual verification. Thus, ensuring strict compliance with the General Data Protection Regulation (GDPR).

As explained in Section \texttt{Spectral calibration}, the raw sensor Digital Numbers (DN) captured by the RGN sensor were converted into calibrated, absolute surface reflectance values. This transformation was executed using the MAPIR Camera Control (MCC) software \cite{mapircameracontrol}. The MCC pipeline ingested the captured images of the diffuse reflectance calibration target acquired at regular intervals to mathematically normalize the RGN imagery.

Additionally, GPS uncertainties were found above a specified level ($>$ 5 meters) for a small number of images. Hence, we labeled the geospatial coordinates for those images as \texttt{NaN} even though the timestamps of data capture can still be used to approximate their location. Finally, the imagery pairs can also be geometrically aligned using the previously derived intrinsic and extrinsic calibration parameters. On a subset of images, the standard stereo-rectification algorithm, implemented in the OpenCV framework \cite{opencv_library}, was applied to establish planar epipolar geometry between the images captured by the different imaging sensors. The scripts for executing this standard geometric rectification, are provided in the repositories mentioned in the Section \texttt{Code Availability}.

The final distribution of images for each morphology is shown in Table \ref{tab:distribution}.

\begin{table}[h!]
\centering
\begin{tabular}{@{}lcc@{}}
\toprule
\textbf{Urban morphology} & \textbf{Total number of images} & \textbf{Images with valid GPS coordinates} \\ \midrule
\textbf{Arnhem (Big urban area)}     & 7,065                 & 6,918       \\
\textbf{Delft (Small city)}      & 3,168                 & 1,851       \\
\textbf{Oirschot (Town)}   & 5,595                 & 4,974       \\
\textbf{Middelbeers (Village)} & 1,890                & 1,833       \\ \addlinespace \midrule

\textbf{Total}       & 17,718               & 15,576       \\ \bottomrule
\end{tabular}
\caption{Distribution of the total number of images for different urban morphologies with valid geospatial location}
\label{tab:distribution}
\end{table}

\subsection*{Use cases}

In this section, we demonstrate a few use cases showcasing the value and utility of this dataset. Each of these use cases can be independent research efforts if analyzed more deeply for the entire dataset. They demonstrate the possibilities enabled by our data collection approach and the dataset itself.

\subsubsection*{Built environment dimension estimation with depth maps using only RGB images without LiDAR}
\begin{figure}[h!]
     \centering
     
     % --- First Row: 3 Images ---
     \begin{subfigure}[h]{0.31\textwidth}
         \centering
         \includegraphics[width=\textwidth]{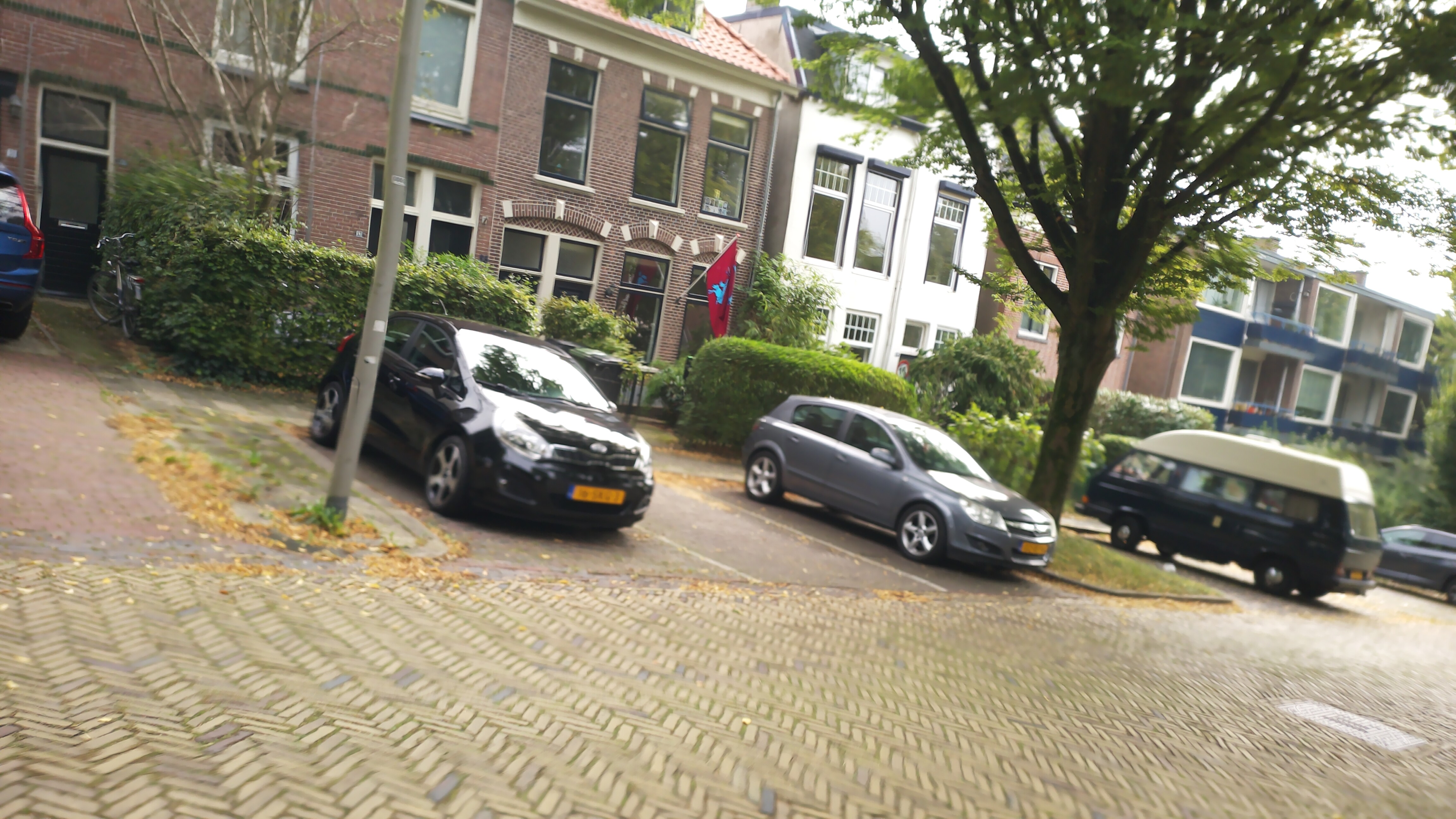}
         \caption{First image used for the depth map generation.}
     \end{subfigure}\hfill
     \begin{subfigure}[h]{0.31\textwidth}
         \centering
         \includegraphics[width=\textwidth]{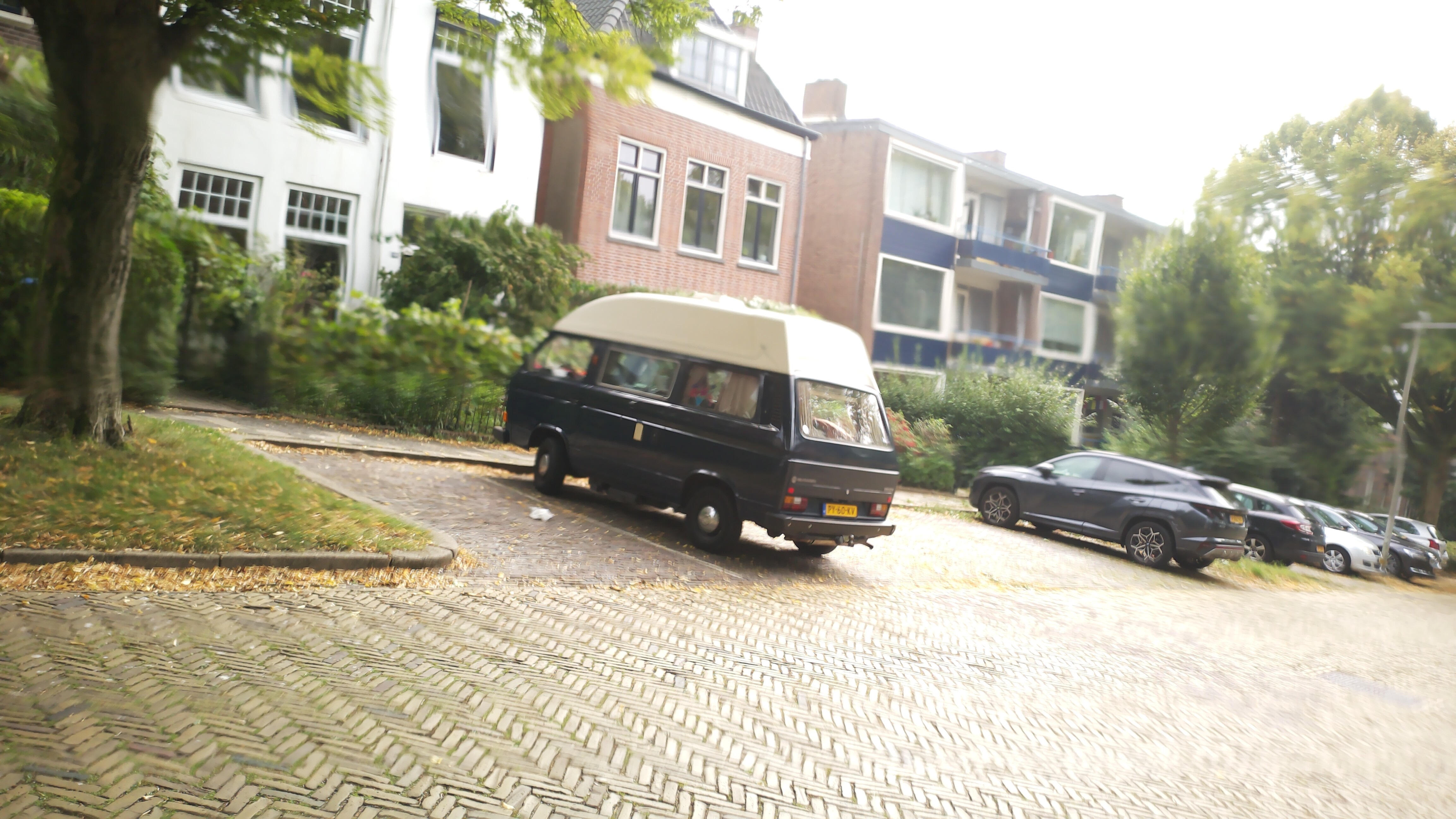}
         \caption{Second image used for the depth map generation.}
     \end{subfigure}\hfill
     \begin{subfigure}[h]{0.31\textwidth}
         \centering
         \includegraphics[width=\textwidth]{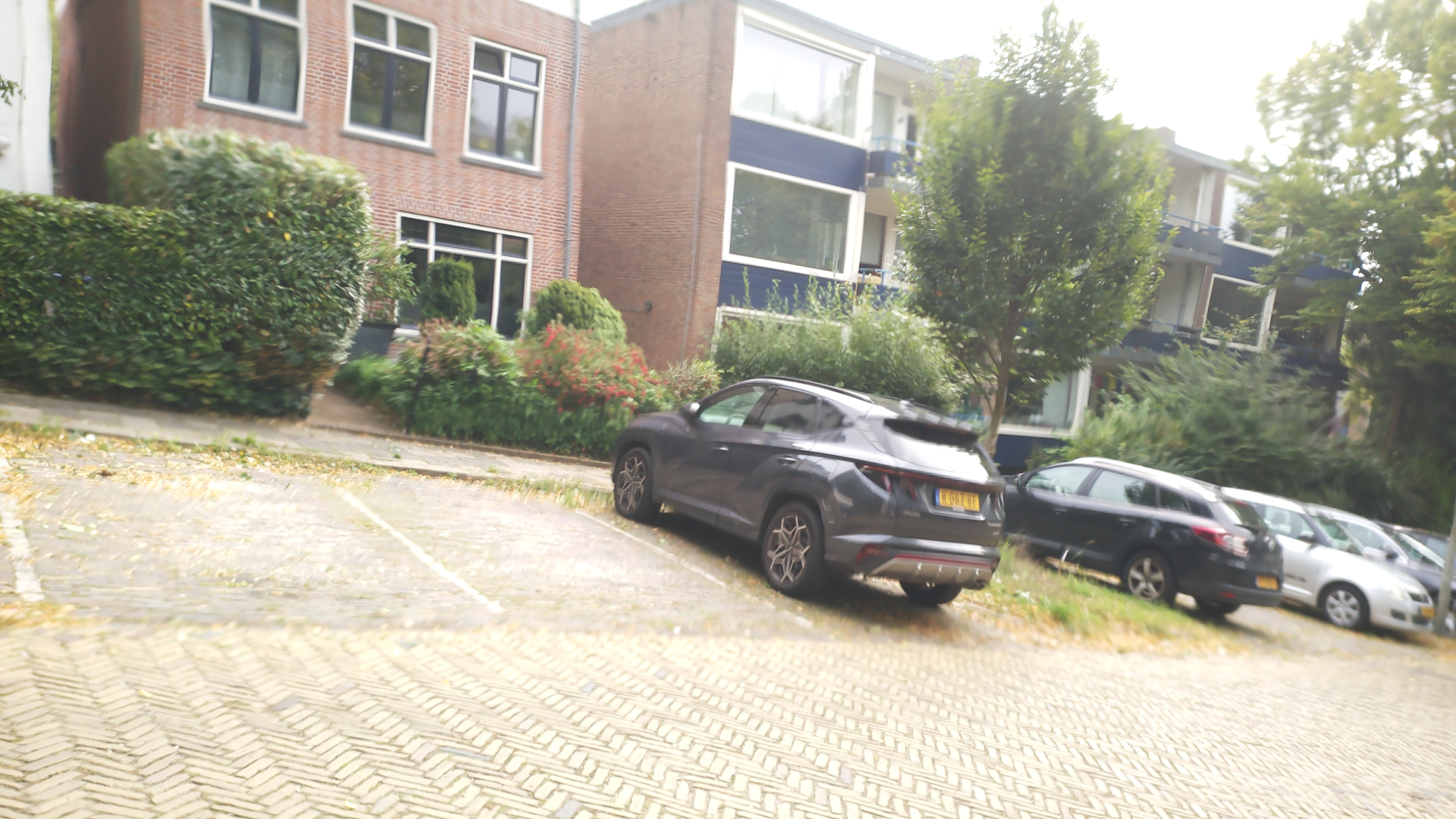}
         \caption{Third image used for the depth map generation.}
     \end{subfigure}
     
     \vspace{15pt} % Adds vertical spacing between rows 1 and 2
     
     % --- Second Row: 2 Images ---
     \begin{subfigure}[h]{0.45\textwidth}
         \centering
         \includegraphics[width=\textwidth]{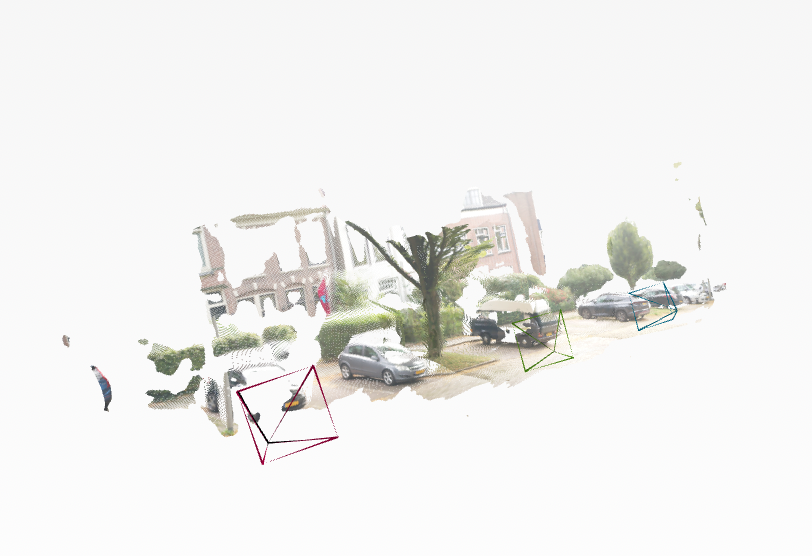}
         \caption{Side view of the generated depth map.}
     \end{subfigure}\hfill
     \begin{subfigure}[h]{0.45\textwidth}
         \centering
         \includegraphics[width=\textwidth]{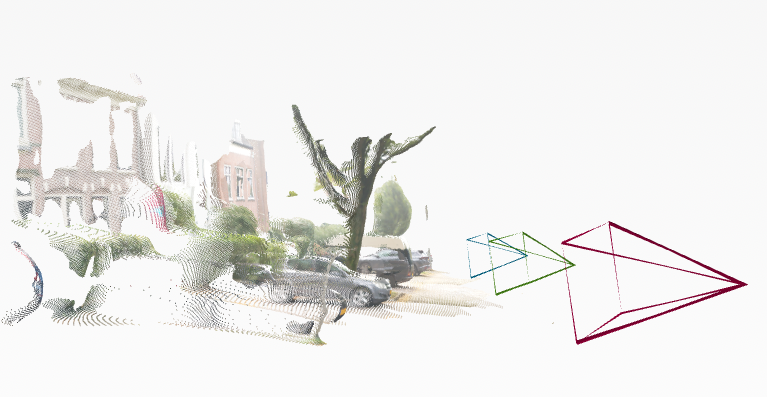}
         \caption{Tangential view of the generated depth map.}
     \end{subfigure}
     
     \vspace{15pt} % Adds vertical spacing between rows 2 and 3
     
     % --- Third Row: 1 Centered Image ---
     \begin{subfigure}[h]{0.45\textwidth}
         \centering
         \includegraphics[width=\textwidth]{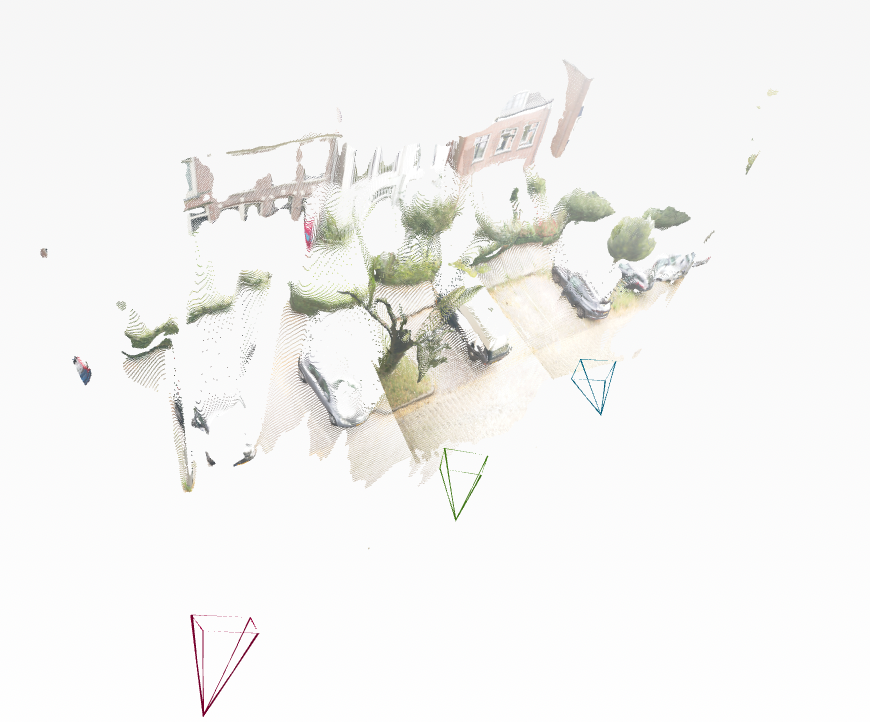}
         \caption{Top view of the generated depth map.}
     \end{subfigure}
    \caption{Depth map of a street in Arnhem using the VGGT model \cite{wang2025vggt}. The red, green, and blue cones denote the estimated positions of the imaging sensor along the bike's trajectory where the images were acquired.}
    \label{fig:depthMap}
\end{figure}

We utilize the state-of-the-art VGGT model \cite{wang2025vggt} for creating a depth map of the entire street using only RGB images captured at fixed intervals. Our approach, paired with the latest advancements in 3D object reconstruction using deep learning, allows the estimation of the dimensions of almost any major object in the field of built environment using just RGB imagery captured at fixed intervals. While the VGGT model \cite{wang2025vggt} and derived approaches have recently become popular for autonomous vehicles and computer vision, they can only generate depth maps with relative dimensions instead of absolute dimensions. Since our metadata contains the exact location of the bike with time-stamps while images are captured, we can estimate the absolute dimensions of the objects in the built environment using the generated depth map by calculating the displacement of the bike using the GPS locations and the time stamp differences. In Figure \ref{fig:depthMap}, we demonstrate the depth map generated using only 3 RGB images captured at fixed intervals for a street in Arnhem. The 3D model file (.GLB) generated for this use-case is provided in the repository mentioned in the Section \texttt{Code Availability}.

This approach can be scaled to entire cities using our data collection approach, to enable the analysis of shadows generated by trees based on the solar position, and effectively calculate the economic value of trees in reducing the Urban Heat Island (UHI) effect. Further, these depth maps can also be combined with airborne LiDAR maps, such as the AHN map of the Netherlands \cite{ahn_dataset}, to generate heights of objects that are not visible from the overhead view.

\subsubsection*{Multi-Spectral Disambiguation of Building Envelopes using RGN and Thermal images}
\label{sec:usecase_disambiguation}

\begin{figure}[htpb]
     \centering
     % --- First Row: 3 Images ---
     \begin{subfigure}[t]{0.32\textwidth}
         \centering
         \includegraphics[width=\textwidth]{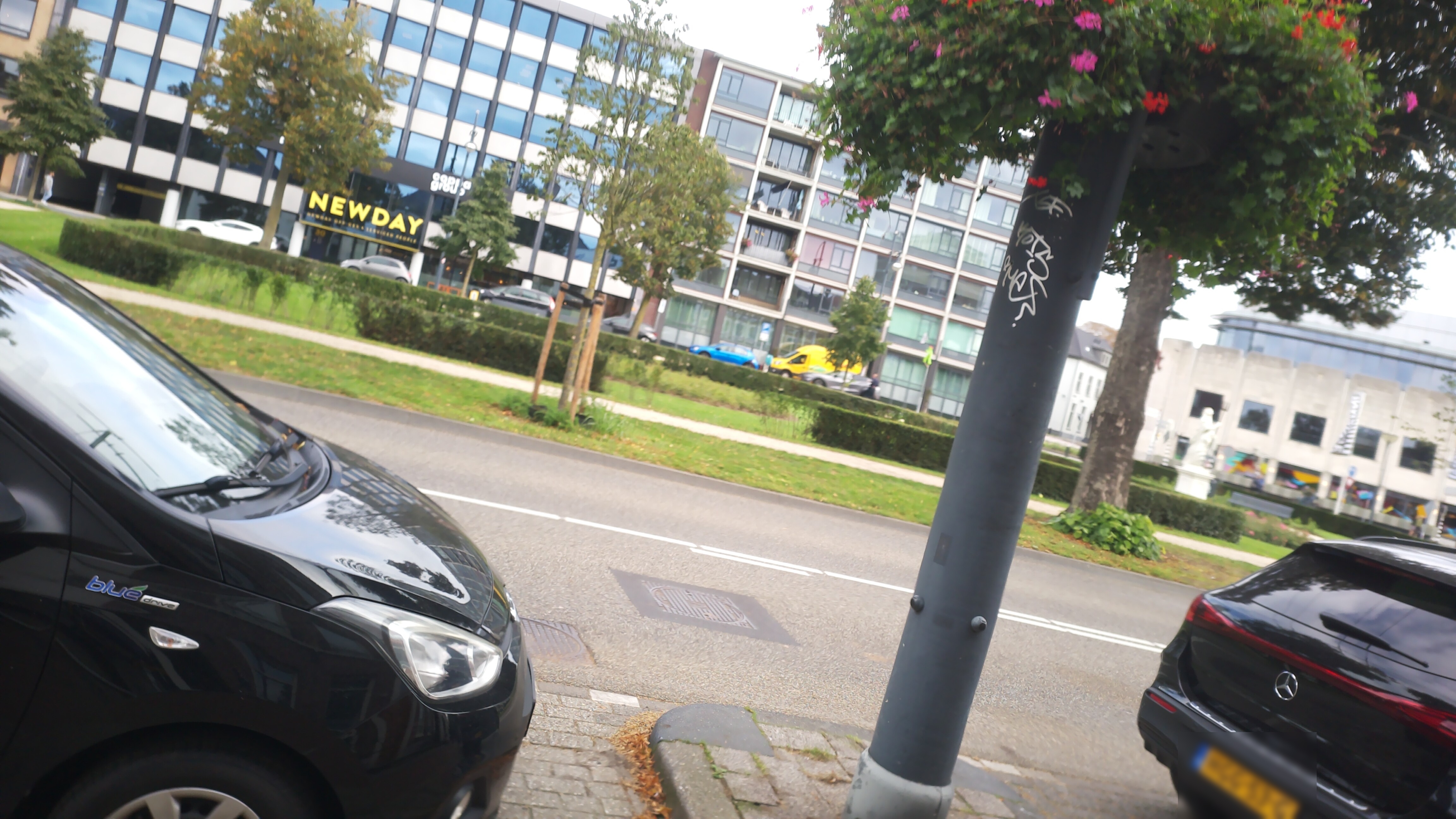}
         \caption{RGB image}
         \label{fig:usecase_rgb}
     \end{subfigure}\hfill
     \begin{subfigure}[t]{0.32\textwidth}
         \centering
         \includegraphics[width=\textwidth]{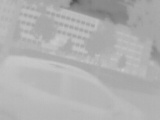}
         \caption{Thermal image}
         \label{fig:usecase_thermal}
     \end{subfigure}\hfill
     \begin{subfigure}[t]{0.32\textwidth}
         \centering
         \includegraphics[width=\textwidth]{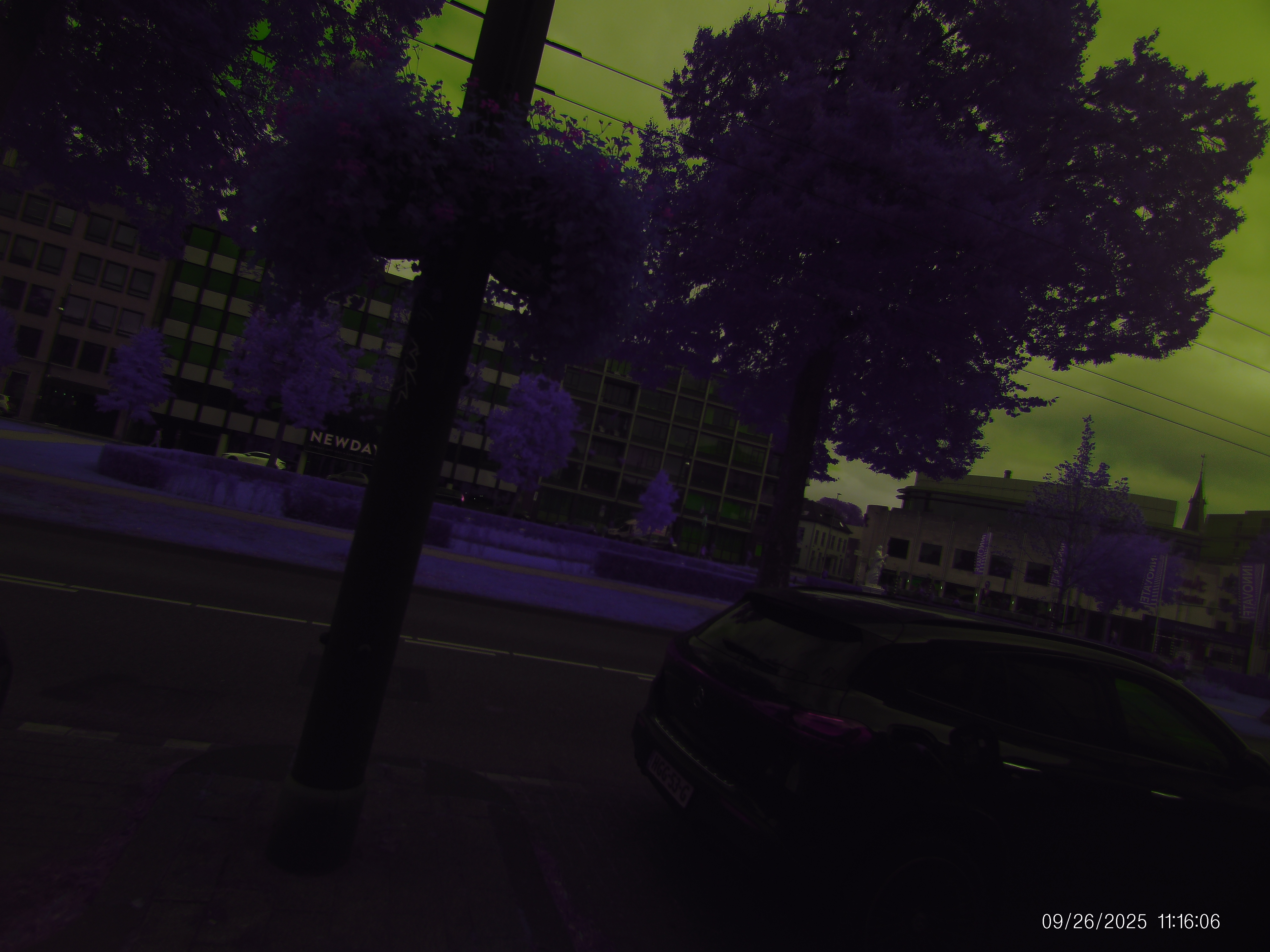}
         \caption{RGN image}
         \label{fig:usecase_rgn}
     \end{subfigure}

     \vspace{10pt} % Adds vertical spacing between rows 1 and 2
     
     % --- Second Row: 3 Images ---
     \begin{subfigure}[t]{0.32\textwidth}
         \centering
         \includegraphics[width=\textwidth]{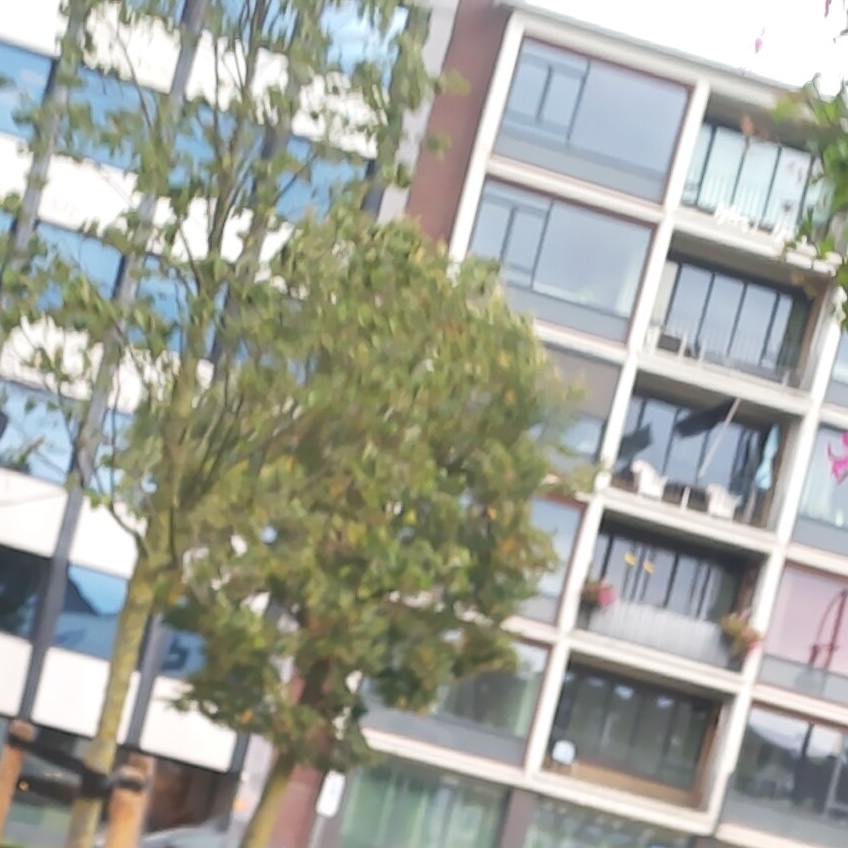}
         \caption{Zoom-in RGB view}
         \label{fig:usecase_rgb_zoomin}
     \end{subfigure}\hfill
     \begin{subfigure}[t]{0.32\textwidth}
         \centering
         \includegraphics[width=\textwidth]{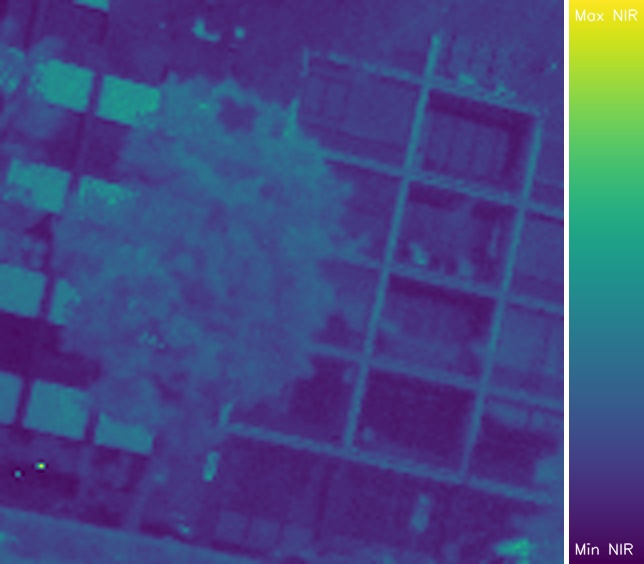}
         \caption{Near-Infrared heatmap}
         \label{fig:usecase_rgn_zoomin}
     \end{subfigure}\hfill
     \begin{subfigure}[t]{0.32\textwidth}
         \centering
         \includegraphics[width=\textwidth]{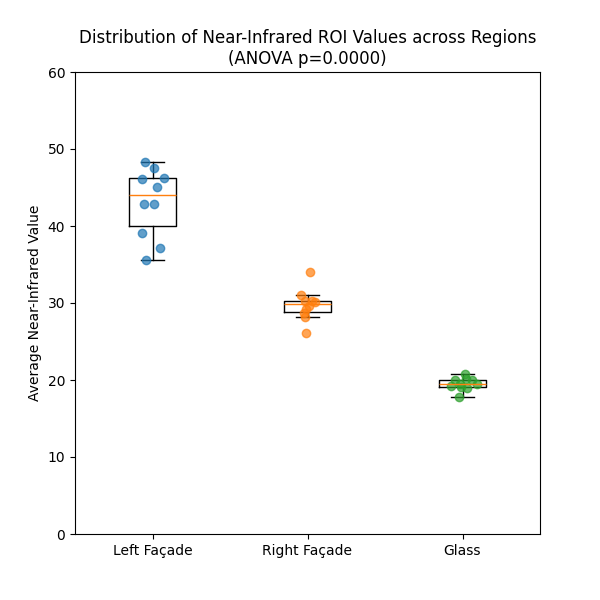}
         \caption{ROI Near-Infrared box-plot}
         \label{fig:usecase_box_plot}
     \end{subfigure}
     \caption{A comparison of building façades from a single vantage point. The near-infrared channel analysis shows that the off-white materials on the two façades, which appear visually similar in the RGB images, are markedly different. Additionally, notice how the temperature of façades are different. \textbf{(a)} Standard RGB (Red, Green, Blue) image with only visible spectra. \textbf{(b)} Thermal image from the same vantage point, where darker pixels indicate lower temperature. \textbf{(c)} RGN (Red, Green, Near-Infrared) image at the same vantage point. \textbf{(d)} Zoom-in view of the RGB image showing how the off-white material on both façades appears visually identical. \textbf{(e)} Zoom-in view of the RGN image with a heat-map applied to the near-infrared channel, demonstrating a marked difference between the off-white materials. \textbf{(f)} Box-plot after selecting 10 different regions-of-interest (ROI)/group of pixels for the off-white material in left façade, the off-white material in right façade and the glass material in the right façade. The y-axis indicates the average near-infrared reflectance for the regions-of-interest in digital format ranging from 0-255.}
    \label{fig:usecase2}
\end{figure}

Street-level imagery has been used extensively to extract information about building envelopes~\cite{2026_autcon_lod3}.
An application of this multi-spectral dataset is a contribution to such a research line -- the disambiguation of urban materials for automated building auditing and thermodynamic modeling. In standard RGB datasets, the different façade materials exhibit spectral similarity, i.e. appearing visually identical due to surface paint, aging, or visual albedo. For instance, several images in this dataset capture scenarios with adjacent building façades that appear to be constructed of an identical material (such as the off-white painted concrete in the RGB imagery as shown in Figures \ref{fig:usecase_rgb} and \ref{fig:usecase_rgb_zoomin}). However, examining the corresponding Near-Infrared spectral bands from the RGN imagery and the Thermal imagery data reveals several material differences and façade divergences.

In the scenario shown in Figure \ref{fig:usecase2}, specifically sub-figures \ref{fig:usecase_rgb} and \ref{fig:usecase_rgb_zoomin}, the façades of both the buildings reflect visible light identically, which will cause standard computer vision segmentation models and recent research works such as \cite{tarkhan2025mapping}, relying on standard computer vision models to classify them as the same building material (e.g. painted concrete).
 
The RGN imagery in Figures \ref{fig:usecase_rgn} and \ref{fig:usecase_rgn_zoomin} reveals a significantly lower near-infrared reflectance for the second façade. This happens because the near-infrared reflectance is highly dependent on chemical composition of the material rather than the superficial color. Further analysis by selecting ten different regions of interest in the off-white material from the two façades is shown in Figure \ref{fig:usecase_box_plot}. This proves that the underlying off-white material on the façades of the adjacent buildings is different.

Finally, the thermal imagery in Figure \ref{fig:usecase_thermal} also shows the differences in the entire façades of the buildings by displaying more darker pixels (lower surface temperature) for the second façade than the first façade. This also correlates with the visually higher glass ratio and indicates a different thermal insulation profile, or building envelope efficiency.

Hence, by fusing these multi-spectral channels, researchers can train advanced segmentation models. Instead of merely classifying a pixel based on the RGB pixel values, machine learning models can be trained on fused multi-spectral vectors for various use-cases including identifying and classifying materials. This enables urban planners and researchers to automate the detection of retrofitted, energy-efficient buildings versus older, heat-retaining structures across a massive urban scale without requiring in-person visits.

%% file: sections/datarecords.tex
\section*{Data Records}

The complete dataset is publicly available and hosted on Zenodo under DOI: 10.5281/zenodo.19440802. The data repository is structured to facilitate seamless integration with standard computer vision, urban planning and photogrammetry pipelines. To ensure temporal fidelity, all image files are strictly named in the order that they were captured based on the creation time of the file. 

The root directory of the repository is organized into three primary sub-directories: \texttt{calibration/} and \texttt{image\_data/}. This is supplemented by a master \texttt{image\_metadata.csv} file.

\subsection*{Calibration Records (\texttt{calibration/})}
This directory contains the files for intrinsic calibration and extrinsic calibration matrices. Further, it contains the image files of the calibration plates, utilized for intrinsic calibration, extrinsic calibration and spectral calibration.

\begin{itemize}
    \item \texttt{intrinsics.json}: Contains the focal lengths ($f_x, f_y$), principal points ($c_x, c_y$), and radial/tangential distortion coefficients for the RGB, RGN, and Thermal imaging sensors.
    \item \texttt{extrinsics.json}: Contains the static $3 \times 3$ rotation matrices ($R$) and $3 \times 1$ translation vectors ($T$) with the RGB sensor designated as the origin $(0,0,0)$ for both RGN and Thermal imaging sensors.
    \item \texttt{calibration\_plates}: Contains images of the calibration plates (black and white checkerboard pattern, modified aluminum tape checkerboard pattern, diffuse reflectance target) utilized for intrinsic, extrinsic, and spectral calibrations.   
\end{itemize}

\subsection*{Image Records (\texttt{image\_data/})}
The imagery data with 17,718 images is structured by the study area of data collection and further structured by the sensor modality, that is, RGB (Red, Green, Blue), RGN (Red, Green, Near-Infrared) and Thermal directories. The name of each file uses a prefix \textbf{studyarea\_123} which indicates the area of the data collection, followed by a number that allows the synchronization of corresponding RGB, RGN, and Thermal imaging files.

\begin{itemize}
    \item \texttt{RGB/}: Contains the RGB imagery ($4608 \times 2592$) saved as 8-bit JPG files.
    \item \texttt{Thermal/}: Contains the hardware-synchronized Thermal imagery ($160 \times 120$). Data is provided as 8-bit JPG files where black values indicates a temperature of 0\textdegree C and white indicates a temperature of 40\textdegree C as the thermal sensor was configured with these temperature values during data collection.
    \item \texttt{RGN/}: Contains the Red, Green, Near-Infrared imagery ($4000 \times 3000$) after spectral calibration processed by the MCC software, saved as 8-bit JPG files.  
\end{itemize}

\subsection*{Metadata Records (\texttt{image\_metadata.csv})}
This file contains the geospatial coordinates (latitude, longitude in WGS84) and timestamps (Coordinated Universal Time UTC time) of data capture for each of the imaging files and the study area. 

Due to licensing restrictions, the Very High Resolution (VHR: 30cm) from the Pleiades NEO satellite imagery as shown in Figure \ref{fig:datasinglePoint} cannot be openly distributed. However, we provide this set of geospatial coordinates which enables researchers with appropriate access to VHR satellite imagery, such as through the ESA Third Party Mission (TPM) \citep{ESA_PleiadesNeo_TPM}, or the Satellite data portal of the Netherlands Space Office (NSO) program, to retrieve identical image tiles from the original provider. The weather data including cloud cover and ambient air temperature during the imagery capture for each of the files can be obtained from the official weather data portal of KNMI (\url{https://dataplatform.knmi.nl/}).

% \subsection*{Raw Data Records (\texttt{raw\_data/})}
% The Red, Green, Near-Infrared imagery ($4000 \times 3000$) after spectral calibration is also provided as 16-bit TIFF files in lossless formats to preserve radiometric and structural integrity. The directory structure is the same as the \texttt{image\_data/} directory. Due to the substantial file size of these uncompressed files, they are hosted in a supplementary, open-access repository via 4TU.ResearchData under DOI: 10.4121/fdb56412-fef0-4d22-91e4-8a0ad07e3269.

%% file: sections/usagenotes.tex
\section*{Usage Notes}

\subsection*{Handling movements of the bike}
While the provided extrinsic matrices accurately describe the physical geometry between the RGB and RGN imaging sensors under static conditions, the effective spatial baseline in the geometric position of the sensors expands dynamically by approximately 0.2 to 0.6 metres depending on the bike pedaling velocity. Furthermore, bike kinematics may have introduced high-frequency micro-rotations (roll and yaw) during the 0.22 second time interval (due to manufacturer constraints on the RGN sensor) between RGB and RGN imagery captures, even though the bikes utilized during data collection had pneumatic tires. To maintain geometric integrity for complex downstream applications such as multi-modal 3D reconstruction, users of this dataset should prefer to combine the provided translation matrices $T$ with dynamic feature-matching algorithms across RGB and RGN images to compensate for both the temporal offset and the non-linear trajectory of the bike platform. Various traditional computer vision techniques and deep learning based methods are available, such as \cite{featureMatching} and \cite{deeplearningbasedVisbileInfrared}.

Additionally, while adding more data by replicating our data collection hardware, the speed of the bike and the shutter speed of the imaging sensors will be related during data collection. Therefore, before data collection, care should be taken in deciding the fixed shutter speed of the imaging sensors based on the maximum speed of the bike (or the moving platform).

\subsection*{Quality control of spectral values}
Imagery data captured with a mobile platform such as bikes, provides several opportunities with advances in terms of spatial and temporal resolutions. However, the imaging sensors utilized in our approach are passive instead of active imaging sensors. Hence, as soon as the conditions of atmosphere change, the assumptions made in \eqref{eq:reflactance}, that made spectral calibration easy to solve for our data collection campaigns, will become more complex. At that point, the spectral values gathered for an object of interest can only be treated as relative values instead of absolute values. Since the relative values for comparing multiple objects of interest might also suffer from atmospheric variations during the data collection campaign, it is crucial to capture repetitive imaging data for the same location if our data collection approach is deployed without using a reflectance calibration target. In general, it is suggested to capture multiple imaging data values across multiple weather conditions and average the relative values to compare similar objects of interest in case of unsupervised data collection without a standard reflectance calibration target. Another approach can be to use atmospheric radiative transfer models such as MODTRAN \cite{1987ssireptRB}, to generate absolute spectral values by using the exact weather conditions as input at the time of imagery data collection.

Additionally, this data set provides a snapshot of a particular location in a particular season or time period. However, measurements in a particular season should not be extrapolated for the entire year or a longer time period. In other words, it is suggested to think carefully whether the environmental variations between seasons for an object of interest will, in turn, produce an eco-physical change in the imaging data for the same object of interest.

Finally, the imaging data captured are limited by the sensitivity of the sensors to their respective spectral band curves and their configuration based on the data collection environment. For example, in our case, the sensors are sensitive to spectral bands as shown in Table \ref{tab:hardware}. The thermal sensor was configured to measure only temperatures between $0^{\circ}\text{C}$ and $40^{\circ}\text{C}$ based on the data collection season.

\subsection*{Future research directions}%  Vision

This dataset directly enables the remote sensing community to calibrate orbital satellite imagery and evaluating atmospheric correction models, by providing high-resolution radiometric baselines. Further, it can also serve as a validation dataset for 3D radiative transfer models \cite{ramiRTM}. For the machine learning and computer vision communities, the spatially registered multi-spectral imagery pairs can serve as a benchmark for image to image translation models (e.g., \cite{pix2pix}). In recent works such as \cite{gupta2025usingstreetviewimagery} and \cite{instantInfrared}, such generative models are increasingly being utilized to synthesize the missing thermal or near-infrared channels from standard RGB street-view imagery datasets for scaling the monitoring of various environmental parameters.

Beyond the immediate dataset, our data collection approach using bikes offers a highly scalable methodology for research requiring high spatio-temporal resolution imagery data. While our data collection campaigns provide a broad spatial coverage, deploying this data collection approach across varying seasons can yield unique insights into dynamic urban micro-climates. For instance, in certain geographies during the winter season, researchers can use this approach, to quantify the liquid water content in snow and predict the formation of hazardous black ice on transit routes \cite{dutchnews2026codered}. Further, future studies could use this methodology to quantify the benefits of blue-green infrastructure, providing street-level validation data for various municipal climate adaptation policies.

%% file: sections/codeAvailablity.tex
\section*{Data Availability}

The dataset generated and analyzed during the current study has been deposited to Zenodo under DOI: 10.5281/zenodo.19440802. Upon formal acceptance of the manuscript, the dataset will be made permanently public under the same reserved identifier 10.5281/zenodo.19440802 with a CC BY 4.0 license.

\section*{Code Availability}
\label{sec:code}

Extensive custom software was written for the development of the hardware platform. The codebase primarily in Python 3.10 and Micropython, manages the synchronized capture of RGB, RGN and Thermal imaging sensors with the GPS coordinates. It relies on standard libraries including \texttt{os}, \texttt{pandas}, \texttt{omv} and \texttt{sensor}. Additional code written in Python 3.12 enabled the generation of intrinsic and extrinsic calibration matrices, the reproduction of use cases and the starter code for stereo alignment. The entire codebase is available on Github at \url{https://github.com/akshitgupta95/urbanScape}. To ensure long-term reproducibility, a released version of the code exactly as used for this research will be archived on Zenodo with the MIT License after acceptance of the manuscript.

%% file: sn-bibliography.bib
@techreport{IPCC,
  author    = {Dodman, D. and Hayward, B. and Pelling, M. and others},
  title     = {Cities, Settlements and Key Infrastructure},
  booktitle = {Climate Change 2022: Impacts, Adaptation and Vulnerability. Contribution of Working Group II to the Sixth Assessment Report of the Intergovernmental Panel on Climate Change},
  publisher = {Cambridge University Press},
  year      = {2022},
  doi       = {10.1017/9781009325844.008},
  url       = {https://www.ipcc.ch/report/ar6/wg2/chapter/chapter-6/}
}

@article{densification1,
AUTHOR = {Dinic Brankovic, Milena and Igic, Milica and Dekic, Jelena and Ljubenovic, Milica},
TITLE = {Impact of Urban Densification on Outdoor Microclimate and Design of Sustainable Public Open Space in Residential Neighborhoods: A Study of {Niš}, {Serbia}},
JOURNAL = {Sustainability},
VOLUME = {17},
YEAR = {2025},
NUMBER = {4},
ARTICLE-NUMBER = {1573},
URL = {https://www.mdpi.com/2071-1050/17/4/1573},
ISSN = {2071-1050},
DOI = {10.3390/su17041573}
}

@article{densification2,
  author  = {Qian, Yun and Chakraborty, T. C. and Li, Jianfeng and Li, Dan and He, Cenlin and Sarangi, Chandan and Chen, Fei and Yang, Xuchao and Leung, L. Ruby},
  title   = {Urbanization Impact on Regional Climate and Extreme Weather: Current Understanding, Uncertainties, and Future Research Directions},
  journal = {Advances in Atmospheric Sciences},
  year    = {2022},
  volume  = {39},
  pages   = {819--860},
  doi     = {10.1007/s00376-021-1371-9}
}

@article{citizenScience,
title = {Advances in portable sensing for urban environments: Understanding cities from a mobility perspective},
journal = {Computers, Environment and Urban Systems},
volume = {88},
pages = {101650},
year = {2021},
issn = {0198-9715},
doi = {https://doi.org/10.1016/j.compenvurbsys.2021.101650},
url = {https://www.sciencedirect.com/science/article/pii/S0198971521000570},
author = {Amit Birenboim and Marco Helbich and Mei-Po Kwan}
}

@article{gupta_2024_tools,
  author = {Gupta, Akshit and Mora, Simone and Preisler, Yakir and Duarte, Fàbio and Prasad, Venkatesha and Ratti, Carlo},
  month = {03},
  pages = {536-544},
  publisher = {Nature Portfolio},
  title = {Tools and methods for monitoring the health of the urban greenery},
  doi = {10.1038/s41893-024-01295-w},
  url = {https://www.nature.com/articles/s41893-024-01295-w#Abs1},
  urldate = {2025-03-04},
  volume = {7},
  year = {2024},
  journal = {Nature Sustainability}
}

@article{desouza_2020_air,
  author = {deSouza, Priyanka and Anjomshoaa, Amin and Duarte, Fabio and Kahn, Ralph and Kumar, Prashant and Ratti, Carlo},
  month = {09},
  pages = {102239},
  title = {Air quality monitoring using mobile low-cost sensors mounted on trash-trucks: Methods development and lessons learned},
  doi = {10.1016/j.scs.2020.102239},
  url = {https://www.sciencedirect.com/science/article/pii/S2210670720304339},
  volume = {60},
  year = {2020},
  journal = {Sustainable Cities and Society}
}

@misc{gupta2025usingstreetviewimagery,
      title={Using street view imagery and deep generative modeling for estimating the health of urban forests}, 
      author={Akshit Gupta and Remko Uijlenhoet},
      year={2025},
      eprint={2504.14583},
      archivePrefix={arXiv},
      primaryClass={cs.CV},
      url={https://arxiv.org/abs/2504.14583}, 
}

@misc{raina2023egoblur,
      title={EgoBlur: Responsible Innovation in Aria},
      author={Nikhil Raina and Guruprasad Somasundaram and Kang Zheng and Sagar Miglani and Steve Saarinen and Jeff Meissner and Mark Schwesinger and Luis Pesqueira and Ishita Prasad and Edward Miller and Prince Gupta and Mingfei Yan and Richard Newcombe and Carl Ren and Omkar M Parkhi},
      year={2023},
      eprint={2308.13093},
      archivePrefix={arXiv},
      primaryClass={cs.CV}
}

@article{greenscan,
author = {Gupta, Akshit and Mora, Simone and Zhang, Fan and Rutten, Martine and Prasad, R. and Ratti, Carlo},
year = {2024},
month = {07},
pages = {1-1},
title = {GreenScan: Toward Large-Scale Terrestrial Monitoring the Health of Urban Trees Using Mobile Sensing},
volume = {PP},
journal = {IEEE Sensors Journal},
doi = {10.1109/JSEN.2024.3397490}
}

@article{tarkhan2025mapping,
  title={Mapping façade materials utilizing zero-shot segmentation for applications in urban microclimate research},
  author={Tarkhan, Nada and Klimenka, Mikita and Fang, Kelly and Duarte, Fabio and Ratti, Carlo and Reinhart, Christoph},
  journal={Scientific Reports},
  volume={15},
  number={1},
  pages={5492},
  year={2025},
  publisher={Nature Portfolio},
  doi={10.1038/s41598-025-86307-1},
  url={https://www.nature.com/articles/s41598-025-86307-1}
}

@article{BRANSON201813,
title = {From Google Maps to a fine-grained catalog of street trees},
journal = {ISPRS Journal of Photogrammetry and Remote Sensing},
volume = {135},
pages = {13-30},
year = {2018},
issn = {0924-2716},
doi = {https://doi.org/10.1016/j.isprsjprs.2017.11.008},
url = {https://www.sciencedirect.com/science/article/pii/S0924271617303453},
author = {Steve Branson and Jan Dirk Wegner and David Hall and Nico Lang and Konrad Schindler and Pietro Perona}
}

@article{2021_trc_bikeability,
 author = {Koichi Ito and Filip Biljecki},
 doi = {10.1016/j.trc.2021.103371},
 journal = {Transportation Research Part C: Emerging Technologies},
 pages = {103371},
 title = {{Assessing bikeability with street view imagery and computer vision}},
 url = {https://www.sciencedirect.com/science/article/pii/S0968090X21003739},
 volume = {132},
 year = {2021}
}

@article{biljecki_2021_street,
  author = {Biljecki, Filip and Ito, Koichi},
  month = {11},
  pages = {104217},
  title = {Street view imagery in urban analytics and {GIS}: A review},
  doi = {10.1016/j.landurbplan.2021.104217},
  volume = {215},
  year = {2021},
  journal = {Landscape and Urban Planning}
}

@article{2024_global_streetscapes,
 author = {Hou, Yujun and Quintana, Matias and Khomiakov, Maxim and Yap, Winston and Ouyang, Jiani and Ito, Koichi and Wang, Zeyu and Zhao, Tianhong and Biljecki, Filip},
 doi = {10.1016/j.isprsjprs.2024.06.023},
 journal = {ISPRS Journal of Photogrammetry and Remote Sensing},
 pages = {216-238},
 title = {Global Streetscapes -- A comprehensive dataset of 10 million street-level images across 688 cities for urban science and analytics},
 volume = {215},
 year = {2024}
}

@article{2026_land_greenery,
 author = {Quintana, Matias and Liu, Fangqi and Torkko, Jussi and Gu, Youlong and Liang, Xiucheng and Hou, Yujun and Ito, Koichi and Zhu, Yihan and Abdelrahman, Mahmoud and Toivonen, Tuuli and Lu, Yi and Biljecki, Filip},
 doi = {10.1016/j.landurbplan.2026.105618},
 journal = {Landscape and Urban Planning},
 pages = {105618},
 title = {It is not always greener on the other side: Greenery perception across demographics and personalities in multiple cities},
 volume = {271},
 year = {2026}
}

@misc{ESA_PleiadesNeo_TPM,
  author       = {{European Space Agency}},
  title        = {{Pléiades Neo ESA Archive}},
  howpublished = {\url{https://earth.esa.int/eogateway/catalog/pleiades-neo-esa-archive}},
  year         = {2026},
  note         = {Accessed: 2026-02-07, Provided via ESA Third Party Mission scheme}
}

@article{USTIN1990293,
title = {Spectral characteristics of ozone-treated conifers},
journal = {Environmental and Experimental Botany},
volume = {30},
number = {3},
pages = {293-308},
year = {1990},
issn = {0098-8472},
doi = {https://doi.org/10.1016/0098-8472(90)90041-2},
url = {https://www.sciencedirect.com/science/article/pii/0098847290900412},
author = {Susan L. Ustin and Brian Curtiss}
}

@techreport{Loudjani2023Pleiades,
  title       = {Report on new {Pl\'{e}iades Neo} sensors benchmark},
  author      = {Loudjani, Philippe and Lemajic, Slavko and Parage, Vincent and Rodrigues, C\'{e}sar},
  year        = {2023},
  institution = {Joint Research Centre (European Commission)},
  type        = {Technical Report},
  number      = {JRC134704},
  doi         = {10.2760/38474},
  address     = {Luxembourg}
}

@manual{RaspberryPiCamera3,
  title        = {Raspberry Pi Camera Module 3 Product Brief},
  author       = {{Raspberry Pi Ltd}},
  year         = {2023},
  url          = {https://www.raspberrypi.com/products/camera-module-3/},
  note         = {Accessed: 2026-01-30}
}

@manual{MapirSurvey3W,
  title        = {Survey3W Camera - Red+Green+NIR (RGN, NDVI) Datasheet},
  author       = {{MAPIR Camera}},
  year         = {2023},
  url          = {https://www.mapir.camera/collections/survey3/products/survey3w-camera-red-green-nir-rgn-ndvi},
  note         = {Accessed: 2026-01-30}
}

@manual{MapirCalibrationPlate,
  title        = {Diffuse Reflectance Standard Calibration Target Package (V2)},
  author       = {{MAPIR Camera}},
  year         = {2023},
  url          = {https://www.mapir.camera/en-gb/products/diffuse-reflectance-standard-calibration-target-package-v2},
  note         = {Accessed: 2026-03-02}
}

@manual{FlirLepton35,
  title        = {Lepton 3.5 Engineering Datasheet (Model 500-0771-01)},
  author       = {{Teledyne FLIR}},
  year         = {2021},
  url          = {https://oem.flir.com/en-hk/products/lepton/?model=500-0771-01&vertical=microcam&segment=oem},
  note         = {Accessed: 2026-01-30}
}

@manual{MicaSenseAltumPT,
  title        = {Altum-PT Multispectral and Thermal Sensor Datasheet},
  author       = {{MicaSense}},
  year         = {2022},
  url          = {https://eaglenxt.com/drone-sensors/altum-pt-camera/},
  note         = {Accessed: 2026-02-01}
}

@manual{FlirBoson640,
  title        = {Boson LWIR Thermal Camera Core Engineering Datasheet (640 x 512, 18mm)},
  author       = {{Teledyne FLIR}},
  year         = {2023},
  url          = {https://www.flircameras.com/product/flir-boson-640x480-18mm-htm/},
  note         = {Accessed: 2026-02-01}
}

@manual{calibCheckerboard,
  title        = {Camera calibration checkerboard target / plate},
  author       = {{calib.io}},
  year         = {2025},
  url          = {https://calib.io/products/checkerboard},
  note         = {Accessed: 2026-03-14}
}

@article{aluminiumcheckerboard,
title = {Systematic approach for thermal imaging camera calibration for machine vision applications},
journal = {Optik},
volume = {247},
pages = {168039},
year = {2021},
issn = {0030-4026},
doi = {https://doi.org/10.1016/j.ijleo.2021.168039},
url = {https://www.sciencedirect.com/science/article/pii/S0030402621015989},
author = {Issac Niwas Swamidoss and Amani {Bin Amro} and Slim Sayadi},
}

@ARTICLE{checkerboardCalibration,
  author={Zhang, Z.},
  journal={IEEE Transactions on Pattern Analysis and Machine Intelligence}, 
  title={A flexible new technique for camera calibration}, 
  year={2000},
  volume={22},
  number={11},
  pages={1330-1334},
  doi={10.1109/34.888718}
  }

@article{opencv_library,
    author = {Bradski, G.},
    journal = {Dr. Dobb's Journal of Software Tools},
    title = {{The OpenCV Library}},
    year = {2000}
}

@techreport{EnMAPFieldGuidesTechnicalReport,
  title       = {Spectral Sampling with the ASD FieldSpec 4 – Theory, Measurement, Problems, Interpretation.},
  author      = {Danner, Martin and Locherer, Matthias and Hank, Tobias and Richter,  Katja},
  year        = {2015},
  institution = {GFZ Data Services},
  type        = {EnMAP Field Guides Technical Report},
  doi         = {10.2312/enmap.2015.008}
}

@article { opportunisticSensingAmsterdam,
      author = "L. W. de Vos and A. M. Droste and M. J. Zander and A. Overeem and H. Leijnse and B. G. Heusinkveld and G. J. Steeneveld and R. Uijlenhoet",
      title = "Hydrometeorological Monitoring Using Opportunistic Sensing Networks in the {Amsterdam} Metropolitan Area",
      journal = "Bulletin of the American Meteorological Society",
      year = "2020",
      publisher = "American Meteorological Society",
      address = "Boston MA, USA",
      volume = "101",
      number = "2",
      doi = "10.1175/BAMS-D-19-0091.1",
      pages=      "E167 - E185",
      url = "https://journals.ametsoc.org/view/journals/bams/101/2/bams-d-19-0091.1.xml"
}

@manual{Airbus2021PleiadesNeo,
  author = {{Airbus Defence and Space}},
  title = {Pléiades Neo: Native 30cm High-Resolution Satellite Imagery},
  institution = {Airbus},
  year = {2021},
  url = {https://space-solutions.airbus.com/imagery/our-optical-and-radar-satellite-imagery/pleiades-neo/},
  note = {Accessed: 2026-03-02}
  
}

@manual{mapircameracontrol,
  author = {{MAPIR}},
  title = {MAPIR Camera Control},
  institution = {MAPIR},
  year = {2024},
  url = {https://mapir.gitbook.io/mapir-camera-control-mcc/interface-tabs/process-tab},
  note = {Accessed: 2026-03-16}
  
}

@misc{wang2025vggt,
  title={VGGT: Visual Geometry Grounded Transformer},
  author={Wang, Jianyuan and Chen, Minghao and Karaev, Nikita and Vedaldi, Andrea and Rupprecht, Christian and Novotny, David},
  booktitle={Proceedings of the IEEE/CVF Conference on Computer Vision and Pattern Recognition},
  year={2025}
}

@misc{ahn_dataset,
  author       = {{Actueel Hoogtebestand Nederland}},
  title        = {Actueel {Hoogtebestand} {Nederland} ({AHN})},
  year         = {2023},
  howpublished = {\url{https://www.ahn.nl/}},
  note         = {Accessed: 2026-03-16}
}

@article{deeplearningbasedVisbileInfrared,
  title={CM-Bench: A Comprehensive Cross-Modal Feature Matching Benchmark Bridging Visible and Infrared Images},
  author={Sun, Liangzheng and He, Mengfan and Shao, Xingyu and Li, Binbin and Yan, Zhiqiang and Li, Chunyu and Meng, Ziyang and Xing, Fei},
  journal={arXiv preprint arXiv:2603.12690},
  year={2026}
}

@article{featureMatching,
      title={XoFTR: Cross-modal Feature Matching Transformer}, 
      author={Önder Tuzcuoğlu and Aybora Köksal and Buğra Sofu and Sinan Kalkan and A. Aydın Alatan},
      year={2024},
      eprint={2404.09692},
      archivePrefix={arXiv},
      primaryClass={cs.CV},
      url={https://arxiv.org/abs/2404.09692}, 
}

@article{ramiRTM,
author = {Widlowski, J.-L. and Pinty, B. and Lopatka, M. and Atzberger, C. and Buzica, D. and Chelle, M. and Disney, M. and Gastellu-Etchegorry, J-P. and Gerboles, M. and Gobron, N. and Grau, E. and Huang, H. and Kallel, A. and Kobayashi, H. and Lewis, P. E. and Qin, W. and Schlerf, M. and Stuckens, J. and Xie, D.},
title = {The fourth radiation transfer model intercomparison (RAMI-IV): Proficiency testing of canopy reflectance models with ISO-13528},
journal = {Journal of Geophysical Research: Atmospheres},
volume = {118},
number = {13},
pages = {6869-6890},
doi = {https://doi.org/10.1002/jgrd.50497},
url = {https://agupubs.onlinelibrary.wiley.com/doi/abs/10.1002/jgrd.50497},
eprint = {https://agupubs.onlinelibrary.wiley.com/doi/pdf/10.1002/jgrd.50497},
year = {2013}
}

@article{pix2pix,
  title={Pix2pix network to estimate agricultural near infrared images from rgb data},
  author={de Lima, Daniel Caio and Saqui, Diego and Mpinda, Steve Ataky Tsham and Saito, Jos{\'e} Hiroki},
  journal={Canadian Journal of Remote Sensing},
  volume={48},
  number={2},
  pages={299--315},
  year={2022},
  publisher={Taylor \& Francis}
}

@article{instantInfrared,
title = {Instant infrared: Estimating urban surface temperatures from street view imagery},
journal = {Building and Environment},
volume = {267},
pages = {112122},
year = {2025},
issn = {0360-1323},
doi = {https://doi.org/10.1016/j.buildenv.2024.112122},
url = {https://www.sciencedirect.com/science/article/pii/S0360132324009648},
author = {Mikita Klimenka and Kevin Zhao and Rainer Hilland and Fan Zhang and James Voogt and Carlo Ratti},

}

@misc{dutchnews2026codered,
  author = {{DutchNews.nl}},
  title = {Code red in the north due to black ice, train services cancelled},
  year = {2026},
  month = {Feb},
  howpublished = {\url{https://www.dutchnews.nl/2026/02/code-red-in-the-north-due-to-black-ice-train-services-cancelled/}},
  note = {Accessed: 2026-03-20}
}

@misc{cbs_keyfigures,
  author       = {Statistics Netherlands},
  title        = {Population dynamics; birth, death and migration per region},
  year         = {2025}, 
  howpublished = {CBS StatLine database},
  url          = {https://opendata.cbs.nl/#/CBS/en/dataset/37259eng/table},
  urldate      = {2026-03-26},
  note         = {Dataset: Population dynamics; birth, death and migration per region. Accessed: 2026-03-26}
}

@article{DENG2021125,
title = {A systematic review of a digital twin city: A new pattern of urban governance toward smart cities},
journal = {Journal of Management Science and Engineering},
volume = {6},
number = {2},
pages = {125-134},
year = {2021},
issn = {2096-2320},
doi = {https://doi.org/10.1016/j.jmse.2021.03.003},
url = {https://www.sciencedirect.com/science/article/pii/S2096232021000238},
author = {Tianhu Deng and Keren Zhang and Zuo-Jun (Max) Shen}
}

@MISC{1987ssireptRB,
author = {{Berk}, Alexander and {Bernstein}, Lawrence S. and {Robertson}, David C.},
title = "{MODTRAN: A moderate resolution model for LOWTRAN}",
howpublished = {Technical Report, 12 May 1986 - 11 May 1987 Spectral Sciences, Inc., Burlington, MA.},
year = 1987,
month = jul,
adsurl = {https://ui.adsabs.harvard.edu/abs/1987ssi..reptR....B}
}

@article{2026_autcon_lod3,
 author = {Ma, Rui and Yang, Chendi and Chen, Jiayu and Biljecki, Filip and Li, Xin},
 doi = {10.1016/j.autcon.2026.106842},
 journal = {Automation in Construction},
 pages = {106842},
 title = {Street view imagery-based method for reconstructing 3D building façade openings},
 volume = {184},
 year = {2026}
}
